\documentclass[10pt,twocolumn,letterpaper]{article}

\usepackage[pagenumbers]{cvpr} 

\usepackage[dvipsnames]{xcolor}
\usepackage{colortbl}
\usepackage[utf8]{inputenc} 
\usepackage[T1]{fontenc}    
\usepackage{url}            
\usepackage{amsfonts}       
\usepackage{nicefrac}       
\usepackage{microtype}      
\usepackage{verbatim}
\usepackage{multirow}
\usepackage{titletoc} 

\usepackage[subtle]{savetrees} 

\usepackage{pifont} 
\newcommand{\cmark}{\ding{51}}
\newcommand{\xmark}{\ding{55}}

\definecolor{aliceblue}{rgb}{0.94, 0.97, 1.0}
\definecolor{t5_red}{rgb}{1.0, 0.97, 0.94}
\definecolor{goodcolor}{rgb}{0.94, 1.0, 0.97} 
\definecolor{mygray}{gray}{0.4}
\newcommand{\testManual}{\textsc{Sent-Test}\xspace}
\newcommand{\valManual}{\textsc{Sent-Val}\xspace}
\newcommand{\testCSLR}{\textsc{CSLR-Test}\xspace}
\newcommand{\modelname}{CSLR$^{2}$\xspace}

\newcommand{\vencsign}{\text{$\mathcal{V}_{enc}^\text{Sign}$}\xspace}
\newcommand{\vencsent}{\text{$\mathcal{V}_{enc}^\text{Sent}$}\xspace}
\newcommand{\tenc}{\text{$\mathcal{T}_{enc}$}\xspace}

\definecolor{retcolordef}{HTML}{0066cc}
\definecolor{clscolordef}{HTML}{ff3399}

\newcommand{\cvprmodifs}[1]{\textcolor{black}{#1}}

\definecolor{cvprblue}{rgb}{0.21,0.49,0.74}
\usepackage[pagebackref,breaklinks,colorlinks,citecolor=cvprblue]{hyperref}

\def\paperID{}
\def\confName{CVPR}
\def\confYear{2024}

\title{A Tale of Two Languages: Large-Vocabulary Continuous \\ \emph{Sign Language} Recognition from \emph{Spoken Language} Supervision}

\author{Charles Raude$^{1,2}$\footnotemark[1]
	\and Prajwal KR$^{2*}$
	\and Liliane Momeni$^{2*}$
	\and Hannah Bull$^{1}$
	\and Samuel Albanie$^{3}$
	\and Andrew Zisserman$^2$
	\and G\"ul Varol$^{1,2}$
	\\
	\vspace{-0.4cm}
	\and
	$^1$ LIGM, \'Ecole des Ponts, Univ Gustave Eiffel, CNRS, France \\
	$^2$ Visual Geometry Group, University of Oxford, UK \\
	$^3$Department of Engineering, University of Cambridge, UK  \\
 {\tt\small \url{https://imagine.enpc.fr/~varolg/cslr2/} }
}

\def\sepappendix{0}

\begin{document}
        \maketitle

	\renewcommand*{\thefootnote}{\fnsymbol{footnote}}
	\footnotetext[1]{Equal contribution.}
	\renewcommand*{\thefootnote}{\arabic{footnote}}

	\begin{abstract}
In this work, our goals are two fold: large-vocabulary continuous sign
language recognition (CSLR), and sign language retrieval. To this end,
we introduce a multi-task Transformer model, \cvprmodifs{\modelname,} that is able to
ingest a signing sequence and output in a \cvprmodifs{joint} 
embedding space
between signed \cvprmodifs{language} and spoken language \cvprmodifs{text}. To \cvprmodifs{enable CSLR evaluation} in the large-vocabulary
setting, we introduce new dataset annotations that
have been manually collected. These provide continuous sign-level
annotations for \cvprmodifs{six hours} of test videos, and will be made
publicly available.
We demonstrate that 
by a careful choice of loss functions, 
training the model for both the CSLR and retrieval tasks is mutually beneficial in terms of performance  -- retrieval improves CSLR performance by providing context, while CSLR improves retrieval with more fine-grained supervision. We further show the benefits of leveraging weak and noisy supervision from large-vocabulary datasets such as BOBSL, namely sign-level pseudo-labels, and English subtitles. 
\cvprmodifs{Our} model significantly outperforms the previous state of the art on both tasks.
\end{abstract}

	\section{Introduction}
\label{sec:intro}

Recognising \textit{continuous} and \textit{large-vocabulary} sign language
is a vital step towards enabling real-world technologies
that enhance communication and accessibility for the deaf or hard of hearing.
With the availability of data that depicts continuous
signing from a large vocabulary of signs~\cite{Albanie21b,camgoz2021content4all,Duarte_CVPR2021},
the computer vision field has recently gained momentum towards this direction, building on previous research that had largely focused on restricted settings such as recognising single signs in isolation~\cite{Joze19msasl,Li19wlasl} or signs covering relatively small vocabularies~\cite{Koller15cslr,Zhou2021ImprovingSL}.

\begin{figure}
    \centering
    \includegraphics[width=1\linewidth]{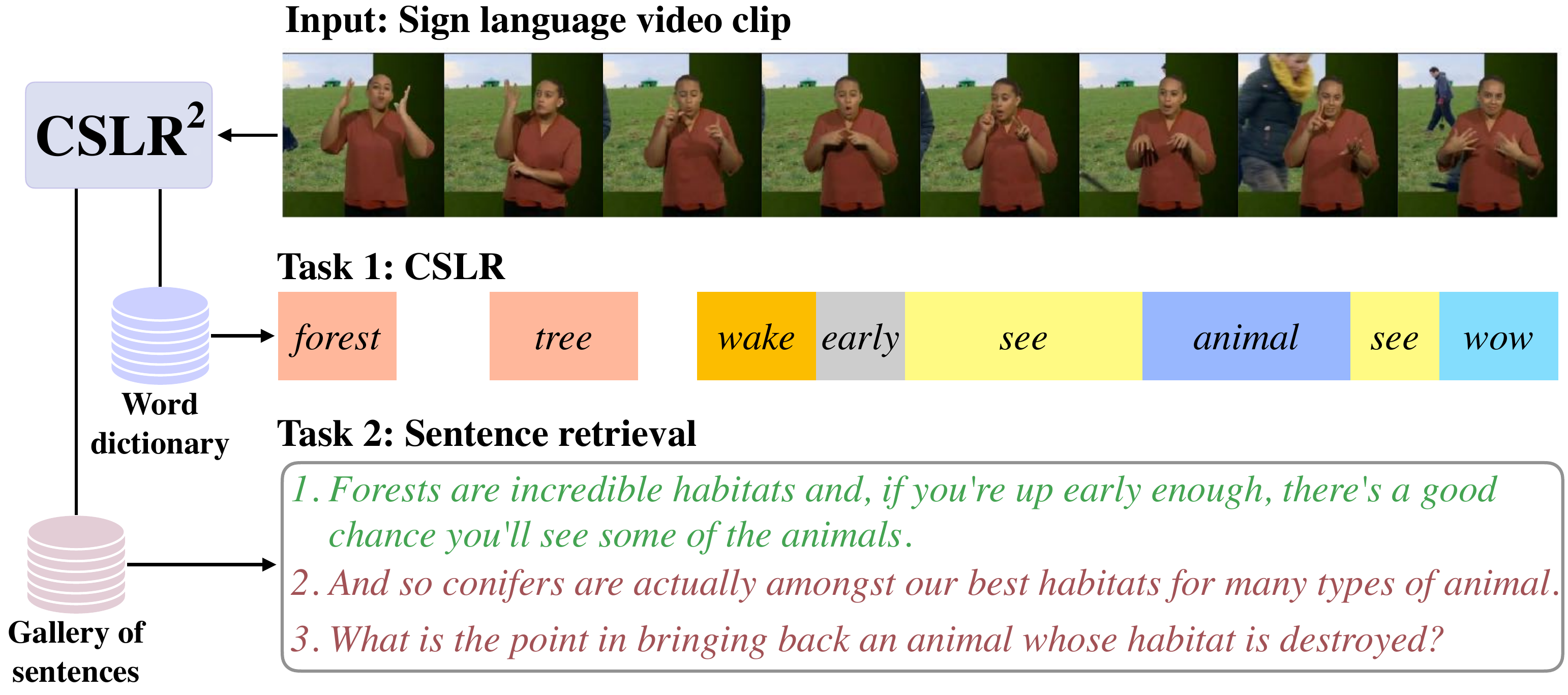}
    \vspace{-0.5cm}
    \caption{
        \textbf{\modelname model}: We illustrate our multi-task model that performs both \textbf{CSLR} and
        sentence \textbf{R}etrieval, thanks to its joint embedding space
        between signed language and spoken language text.
    }
    \label{fig:teaser}
    \vspace{-0.4cm}
\end{figure}

Our goal in this paper is two-fold: first, to enable \textit{large-vocabulary}
continuous sign language recognition (CSLR) \cvprmodifs{-- providing time aligned and dense word predictions for each sign within a signing sequence.} This is an essential first step towards translation, 
as English sentence-level annotations have been shown to be difficult to use
directly as targets for sign language translation~\cite{Varol21,Albanie21b,camgoz2021content4all}.
Our second goal
is {\em \cvprmodifs{sentence} retrieval}, i.e.~\cvprmodifs{given a signing video, to retrieve the most similar sentence text or vice versa} (see Fig.~\ref{fig:teaser}).
This is important as indexing sign language videos
to make them searchable has been highlighted as a useful application for deaf or hard of hearing~\cite{bragg2019}. Also, video to subtitle retrieval can be seen as a proxy for translation -- it is reminiscent of the 
pre-deep learning style of machine translation where sentences were broken down into phrases
and translation proceeded by a lookup of paired phrases in the two languages~\cite{koehn-etal-2003-statistical}.

There are several challenges to achieving these goals, primarily due to the lack of suitable data for 
training and evaluation.
For CSLR, ideally, \textit{each} individual sign within a continuous video
should be associated to a symbolic category. However, current training supervision sources are restricted by their \textit{weak} or \textit{sparse} nature.
For instance, in the largest dataset BOBSL~\cite{Albanie21b}, the available annotations 
are either (i) at sentence-level, weakly associating the entire signing video to an English sentence,
rather than breaking the video into individual sign-word correspondences,
or (ii) at sign-level, but sparse with gaps in the temporal timeline (despite the densification
efforts in~\cite{Momeni22} to scale up the number and vocabulary of annotations). Also, there is no evaluation benchmark with continuous ground truth sign annotations for the BOBSL dataset, so it is
not possible to assess and compare the performance of large-vocabulary CSLR algorithms at scale.

In this paper, we introduce a simple Transformer encoder model~\cite{Vaswani2017} that ingests a signing video sequence and outputs
tokens in a \cvprmodifs{joint} 
embedding space between signed and
spoken\footnote{\cvprmodifs{We refer to the written 
form of spoken language, not the speech audio.}}
languages. The output space enables both the
CSLR and \cvprmodifs{sentence} retrieval tasks.
The Transformer architecture outputs CSLR predictions by leveraging temporal context, and a retrieval embedding through pooling.
The \cvprmodifs{joint} 
embedding language
space \cvprmodifs{may also help} 
to overcome
yet another challenge of sign language recognition: polysemy
where the same word may correspond to several sign variants,  and conversely the same sign may
correspond to several different words.

We train our model on both tasks by leveraging noisy supervision from the large-scale BOBSL dataset. Specifically, we use an individual sign predictor to generate continuous pseudo-labels (for training CSLR) and available weakly-aligned sentence-level 
annotations (for training \cvprmodifs{sentence} retrieval). We show that training for both tasks is mutually beneficial -- in that
including CSLR improves the retrieval performance, and including retrieval improves the CSLR performance. To enable CSLR evaluation, we manually collect new sign-level annotations that are continuous
on the timeline. Since we focus on \cvprmodifs{the large-vocabulary setting, we collect annotations} on the BOBSL test set.
\cvprmodifs{We hope our new CSLR benchmark will facilitate further exploration in this field.}

In summary, our contributions are the following:
(i) We demonstrate the advantages of a single
model, \modelname, that is trained jointly for
both CSLR and sign language sentence retrieval with weak supervision.
(ii) Thanks to our \cvprmodifs{joint} 
embedding space between spoken and
signed languages, we are the first to perform sign recognition
via video-to-text retrieval.
(iii) We build a 
benchmark \cvprmodifs{of substantial size} for evaluating \cvprmodifs{large-vocabulary}
CSLR
by collecting continuous sign-level annotations for \cvprmodifs{6 hours} of video.~(iv) We significantly outperform strong baselines on our new CSLR
and retrieval benchmarks, and carefully ablate each of our components.
We make our code and data available for research.

	\section{Related Work}
\label{sec:related}

We briefly discuss
relevant works
that operate on (i)~continuous sign language
video streams, (ii)~sign language retrieval, and (iii)~CSLR benchmarks.

\begin{table}
	\centering
	\resizebox{1\linewidth}{!}{
		\begin{tabular}{clcccccc}
			\toprule
			& & segmented & \#sentences & hours & vocab. & \#glosses & source \\
			\midrule
			\multirow{3}{*}{train} & PHOENIX-2014~\cite{Koller15cslr} & \xmark & \textcolor{white}{00}6K & \textcolor{white}{00}11 & 1K & \textcolor{white}{0}65K\textcolor{white}{*} & TV \\
			& CSL-Daily~\cite{Zhou2021ImprovingSL} & \xmark & \textcolor{white}{0}18K & \textcolor{white}{00}21 & 2K & 134K\textcolor{white}{*} & lab \\
    			& BOBSL~\cite{Albanie21b} & \xmark & 993K & 1220 & \cvprmodifs{72K*} & \textcolor{white}{..}\cvprmodifs{5.5M*} & TV \\
			\midrule \midrule
			\multirow{3}{*}{test} & PHOENIX-2014~\cite{Koller15cslr} & \xmark & \textcolor{white}{0}629 & 1.0 & 0.5K\textcolor{white}{*} & \textcolor{white}{0}7.1K\textcolor{white}{*} & TV \\
			& CSL-Daily~\cite{Zhou2021ImprovingSL} & \xmark & 1176 & 1.4 & 1.3K\textcolor{white}{*} & \textcolor{white}{0}9.0K\textcolor{white}{*} & lab \\
		\rowcolor{aliceblue} & BOBSL \testCSLR & \cmark & 4518 & 6.0 & 5.1K\textcolor{white}{*} & 32.4K\textcolor{white}{*} & TV \\
	
       \bottomrule
		\end{tabular}
	}
	\caption{\textbf{Recent CSLR training and evaluation sets:}
	Our manually-curated BOBSL \testCSLR set is larger in number of annotated signs and vocabulary, compared to other CSLR test sets from the literature. In addition, it also comes with sign segmentation annotations.
	\cvprmodifs{*Note that the BOBSL training has different vocabulary sets of varying sizes: 72K words spanned by subtitles and 25K words spanned by 5.5M automatic annotations generated in~\cite{Momeni22}.}
	}
	\label{tab:datasets}
	\vspace{-0.4cm}
\end{table}

\noindent\textbf{Ingesting continuous sign language video streams.}
In the recent years, the community has started to move 
beyond isolated sign language recognition (ISLR)~\cite{Joze19msasl,Li19wlasl},
which only seeks to assign a category (typically also
expressed as a word) to a short video segment trimmed
around a single sign without context.
Besides CSLR, several tasks that require ingesting a continuous video stream exist. These include
sign spotting~\cite{Albanie20,Momeni20b,Varol21,Momeni22},
sign tokenization~\cite{Renz21a,Renz21b},
translation~\cite{Camgoz2018,camgoz2020sign,camgoz2020multichannel,zhou2023gloss,yin2023gloss},
subtitle alignment~\cite{Bull21}, subtitle segmentation~\cite{bull2020},
text-based retrieval~\cite{duarte22slretrieval,cheng2023cico}, fingerspelling detection~\cite{Prajwal22a},
active signer detection and diarization~\cite{Albanie21a}.
Our work is related to some of these works
in that they also operate on a large-vocabulary setting
\cite{Albanie20,Momeni20b,Varol21,Bull21,Momeni22};
however, they do not tackle CSLR, mainly due to lack of continuous sign annotations.
While~\cite{duarte22slretrieval} addresses retrieval, their method is not suitable for CSLR -- our work differs in that we perform
both tasks jointly.

State-of-the-art CSLR methods have so far 
focused on PHOENIX-2014~\cite{Koller15cslr} or CSL-Daily~\cite{Zhou2021ImprovingSL}
benchmarks, where the performances are saturated. 
These methods typically consider a fully-supervised setting,
and train with RNN-based~\cite{Huang2018VideobasedSL,Cui2019,zhang2023c2st},
or Transformer-based~\cite{camgoz2020sign} models.
Due to lack of sign segmentation annotation (i.e., the start and end times of signs are unknown),
many works use the CTC loss~\cite{camgoz2020sign,cheng2020,jiao2023cosign,wei2023improving,zuo2022c2slr}.
Our work differs from these previous works on several fronts. We consider a weakly-supervised
setting, where the training videos are \emph{not} annotated for CSLR purposes,
but are accompanied with weakly-aligned spoken language translation sentences. We also study the benefits of joint training with CSLR and retrieval objectives. 
\cvprmodifs{In a similar spirit, the works of \cite{camgoz2020sign,zuo2022c2slr}
jointly train CSLR with sentence-level objectives (translation in \cite{camgoz2020sign}, margin loss for gloss-sequence text retrieval in \cite{zuo2022c2slr}), but in significantly different settings (e.g.,~$8\times$ smaller vocabulary, and with manually-annotated CSLR labels for training).} 

\noindent\textbf{Sign language retrieval.} 
Early works focused on query-by-example~\cite{athitsos2010large,zhang2010using}, where the goal is to retrieve individual sign instances for given sign examples.
The release of continuous sign video datasets, 
like BOBSL~\cite{Albanie21b}, How2Sign~\cite{Duarte_CVPR2021}, and CSL-Daily~\cite{Zhou2021ImprovingSL} with (approximately) aligned spoken language subtitles, 
has shifted the interest towards spoken language to sign language retrieval (and vice-versa).
The first work in this direction is the recent method of~\cite{duarte22slretrieval}, which focuses on improving
the video backbone that is subsequently used for a simple retrieval model using a contrastive margin loss.~CiCo~\cite{cheng2023cico} also focuses on improving video representations, specifically, by
designing a domain-aware backbone.
In contrast, our main emphasis is on (i) the use of weakly-supervised data,
and (ii) the joint training with CSLR. 

Our work naturally derives lessons from the large number of efforts in the parent task of sign language retrieval, i.e., video-text retrieval~\cite{Bain21,gabeur2020multi,liu2021hit,Liu19a,luo2022clip4clip,sun2019videobert,yu2018joint}. Works such as CoCa~\cite{yu2022coca} and JSFusion~\cite{yu2018joint} have shown that jointly training with a cross-modal retrieval objective can help in other tasks such as captioning and question-answering. 
Our approach 
is in the same vein as these works:
we show that jointly training for retrieval and CSLR improves performance for both tasks.

\noindent\textbf{CSLR benchmarks.}
Early works with continuous signing videos provided very small
vocabularies in the order of several hundreds (104 signs in Purdue RVL-SLLL~\cite{purdue06} and
BOSTON104~\cite{EfficientApproxJointTrackingRecognition} ASL datasets,
178 in CCSL~\cite{Huang2018VideobasedSL},
310 in GSL~\cite{adaloglou2020comprehensive},
455 in the SIGNUM DGS dataset~\cite{signum2008},
and 524 in the KETI KSL dataset~\cite{ko2019neural}).
BSL Corpus~\cite{schembri2013building} represents
a large-vocabulary collection; however, it is mainly curated for
linguistics studies, and has not been used for CSLR.

Relatively large collections made it possible to
train CSLR methods based on neural networks (see Tab.~\ref{tab:datasets}).
Most widely used RWTH-PHOENIX-Weather 2014~\cite{Koller15cslr} dataset
contains around 11 hours of videos sourced from weather forecast on TV.
CSL-Daily~\cite{Zhou2021ImprovingSL} provides 20K videos with gloss and translation
annotations from daily life topics, covering a 2K sign vocabulary, and 23 hours of lab recordings.
In Tab.~\ref{tab:datasets}, we provide several statistics 
to compare against our new CSLR benchmark, mainly on their evaluation sets (bottom).
While being larger, we also provide sign segmentation annotations.

Recently released large-vocabulary continuous datasets
(such as
BOBSL~\cite{Albanie21b},
How2Sign~\cite{Duarte_CVPR2021},
Content4All~\cite{camgoz2021content4all}, and
OpenASL~\cite{shi2022open})
do not provide sign-level gloss annotations
due to the prohibitive costs of densely labeling within the open-vocabulary setting.
In this work, we leverage an isolated sign recognition model to generate continuous pseudo-labels for training CSLR, and available weakly-aligned sentence-level supervision for retrieval.

	\begin{figure*}
	\centering
	\includegraphics[width=1\linewidth]{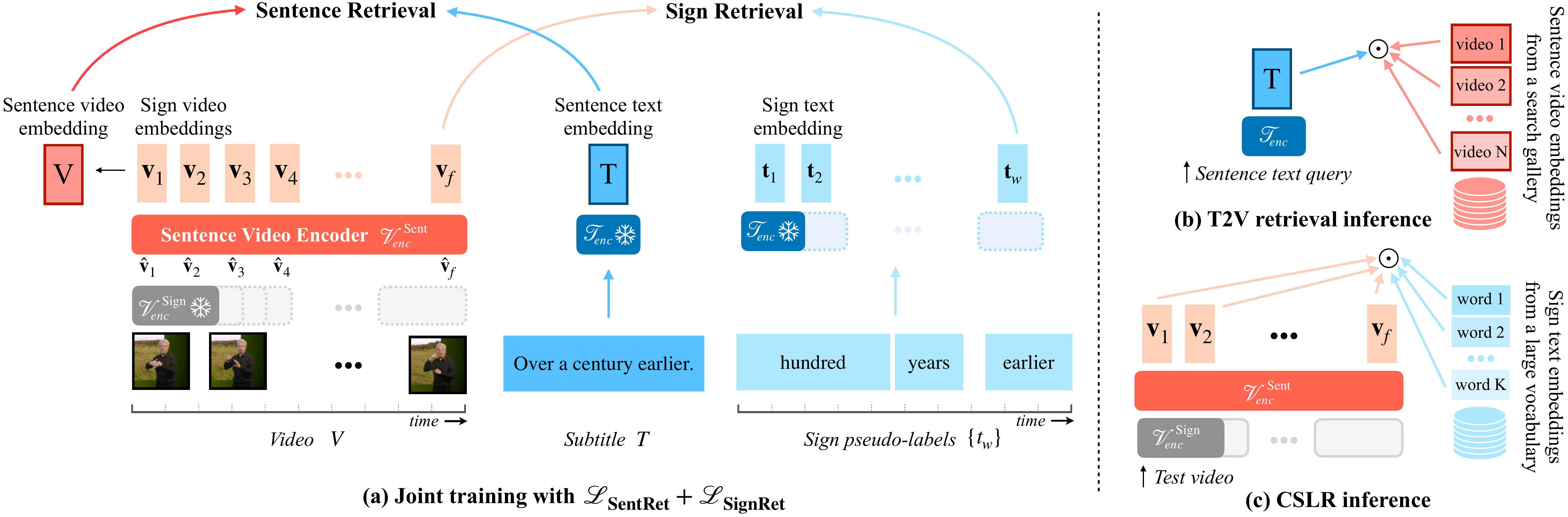}
	\vspace{-0.7cm}
	\caption{
		\cvprmodifs{\textbf{Method overview:}
			(a)~We show a simplified view for our model architecture which consists of both video and text streams.
			On the video side, features are extracted from a signing video clip $V$ by running \vencsign in a sliding 
			window fashion and passed through a Transformer model \vencsent.
			A video embedding $\mathbf{V}$ and sign video embeddings $\{\mathbf{v}_f\}$ are subsequently extracted.
			On the text side, we input an English subtitle sentence $T$ and sign pseudo-labels 
			$\{t_w\}$ to the text encoder \tenc and obtain sentence and sign text 
			embeddings ($\mathbf{T}$, $\mathbf{\{\mathbf{t}_w\}}$), respectively.
			While we illustrate only one triplet data point $(V, T, \{t_w\})$, in practice, we operate on a minibatch of triplets,
			and employ two contrastive losses to jointly train on sentence retrieval $\mathcal{L}_{\text{SentRet}}$ and sign retrieval $\mathcal{L}_{\text{SignRet}}$.
			(b)~For text-to-video retrieval inference, we simply extract a sentence text embedding
			given a text query, and rank the sentence video embeddings corresponding to gallery videos according to their cosine similarities.
			(c)~For CSLR inference, each sign video embedding is matched to the top-ranked word from a large vocabulary of size 8K. A post-processing strategy is applied on frame-level predictions to produce final outputs. For visibility, we omit linear layers which project embeddings into the learnt joint-space. See Sec.~\ref{subsec:architecture} for a detailed description of the architecture and inference procedure. 
        }
    }
    \label{fig:method}
    \vspace{-0.4cm}
\end{figure*}

\section{Joint Space for Signed and Spoken Languages}
\label{sec:method}

We start by describing the model design that goes from raw sign language video pixels
to a {\em joint embedding space} with spoken language text (Sec.~\ref{subsec:architecture}). We then present the losses of our joint training framework with sentence-level and sign-level objectives (Sec.~\ref{subsec:objectives}).
Next, we detail our supervision which consists of
(noisy) sign-level pseudo-labels and weakly-aligned subtitles (Sec.~\ref{subsec:supervision}).
Finally, we provide model implementation details (Sec.~\ref{subsec:implementation_details}).

\subsection{Model overview and inference}
\label{subsec:architecture}

Our model, shown in Fig.~\ref{fig:method}, consists of three main components: 
(i) a sign video encoder \vencsign based on a pretrained Video-Swin~\cite{Liu2022VideoST},
(ii) a sentence video encoder \vencsent based on a randomly initialised Transformer encoder,
and
(iii) a text encoder \tenc based on a \cvprmodifs{pretrained T5 model~\cite{raffel2020t5}}.
Given raw RGB video frame pixels $V$ for a signing sentence,
we obtain a sequence of isolated sign video embeddings from \vencsign as $\{\mathbf{\hat{v}}_f\} = \vencsign(V)$. 
In practice, such a sequence is obtained by feeding 16 consecutive frames to the sign video encoder in a sliding window fashion, with a stride of 2 frames.
These are then fed through the sentence video encoder \vencsent
to have \emph{context-aware} sign video embeddings $\{\mathbf{v}_f\}$, as well as 
a single sentence video embedding $\mathbf{V}$,
denoted
$(\{\mathbf{v}_f\}, \mathbf{V}) = \vencsent(\{\mathbf{\hat{v}}_f\})$.
Similarly, for the text side, we embed the sentence $T$ into $\mathbf{T} = \tenc(T)$.
Additionally, we define sign-level text embeddings for each sign in the sentence as $\{\mathbf{t}_w\}_{w=1}^W = \{\tenc(t_w)\}_{w=1}^W$,
where $W$ is the number of signs in the sentence. 
In practice, we get these embeddings by independently feeding the word(s) corresponding to each sign to the text encoder.

\noindent\textbf{CSLR inference.}
Sign-level recognition predictions are obtained by using 
the sequence of sign video embeddings in $\mathbf{v}_f$ that lies in the 
same space as spoken language.
To associate each feature frame $f$ to a word (or phrase), we perform nearest neighbour classification
by using a large text gallery of sign category names, as illustrated in 
Fig.~\ref{fig:method}\textcolor{red}{c}.
In our experiments, we observe superior performance of such retrieval-based classification
over the more traditional cross-entropy classification \cite{wang2021actionclip},
with the advantage that it is potentially not limited to a closed vocabulary.
In order to go from per-feature classification to continuous sign predictions, we perform a post-processing strategy detailed in Sec.~\ref{subsec:supervision}. We note that this same post-processing strategy is used for obtaining our sign-level pseudo-labels for training.

\noindent\textbf{Retrieval inference.}
For sign-video-to-text (V2T) retrieval, the video sentence embedding $\mathbf{V}$ is matched to a gallery of text sentence embeddings, ranking text sentences by their cosine similarities. Symmetrically, text-to-sign-video (T2V) retrieval is performed in a similar manner, as shown in Fig.~\ref{fig:method}\textcolor{red}{b}.

\subsection{Training with sentence- and sign-level losses}
\label{subsec:objectives}
We train the Transformer-based model, that operates on sentence-level sign language
videos,
to perform two tasks, namely, \cvprmodifs{CSLR and sign language Retrieval (\modelname).}
As illustrated in Fig.~\ref{fig:method}\textcolor{red}{a}, we employ two retrieval losses:
(i) a sentence-level objective, supervised with weakly-aligned subtitles, and 
(ii) a sign-level objective, supervised with pseudo-labels obtained from a strong ISLR model~\cite{Prajwal22a}.
We next formulate each objective individually before introducing our joint framework
that leverages
both sentence-level and sign-level information. 

\noindent\textbf{Sentence-level objective: sign language sentence retrieval (SentRet).}
We explore the task of retrieval as a means to obtain supervision signal
from the subtitles.
Following the success of vision-language models building
a cross-modal embedding between images and text \cite{clip2021,BLIP2022}, we employ a standard contrastive loss, and map sign language videos to spoken language text space.

Sign language sentence retrieval is made of two symmetric tasks, that is, V2T and T2V retrievals. 
For the former, given a query \cvprmodifs{signing} video $V$, the goal is to rank a gallery of text samples (here subtitles)
such that the content of $V$ matches the content in the highest ranked texts. 
Symmetrically, in the latter, given a text query $T$, the goal is to rank a gallery of signing videos.

Formally, given a dataset $\mathcal{D} = \{(V_i, T_i)\}_{i = 1}^N$ of video-subtitle pairs, the goal is to learn two encoders  $\phi_{V}, \phi_{T}$ mapping each signing video $V$ and subtitle $T$ into a joint embedding space. In the following, $\mathbf{V}_i = \phi_{V}(V_i)$ and $\mathbf{T}_i = \phi_{T}(T_i)$ denote the video and text embeddings, respectively.
The encoders are trained using a recently proposed Hard-Negative variant of InfoNCE~\cite{infonce}, HN-NCE~\cite{hnnce2023},
that re-weighs the contribution of each element in the computation of the contrastive loss.
Let $\{(\mathbf{V}_i, \mathbf{T}_i)\}_{i = 1}^B$ be a batch of encoded video-subtitle pairs
and $S_{ij} = \mathbf{V}_i^T \mathbf{T}_j$ be the similarity between the pair $(i,j)$.
For the sake of visibility, we only detail the equations for V2T:
\begin{equation}
    \label{eq:hnnce}
    \mathcal{L}_{\text{HN-NCE}, \text{V2T}} = 
     - \dfrac{1}{B} \sum_{i = 1}^B \log \dfrac{e^{S_{ii} / \tau }}{\alpha \cdot e^{S_{ii} / \tau}
     + \sum_{j \neq i} w^{V2T}_{ij} \cdot e^{S_{ij} / \tau} },
\end{equation}
with $w^{V2T}_{ij}$ weights defined as
\begin{equation}
    \label{eq:weights_hnnce}
    w^{V2T}_{ij} = \dfrac{(B - 1) \cdot e^{\beta S_{ij} / \tau}}{\sum_{k \neq i} e^{\beta S_{ik} / \tau}},
\end{equation}
where the temperature $\tau > 0$, $\alpha \in (0, 1]$, and $\beta \geq 0$ are hyperparameters.
\cvprmodifs{By training to maximise the similarity between correct pairs of video 
and subtitle embeddings, while minimising the similarity between negative pairs,
HN-NCE serves as a proxy for the retrieval by ranking that we perform at inference.}

\noindent\textbf{Sign-level objective: sign 
classification via sign retrieval (SignRet).}
Given a continuous signing video, the goal of CSLR is to recognise a sequence of individual signs.
The continuous video is encoded into a sequence of context-aware sign video embeddings $\{\mathbf{v}_f\}_{f=1}^F$,
with $F$ the number of video frames. 
Again, since these embeddings are in the same joint space as the text embeddings, 
a contrastive loss can be used as a proxy for sign retrieval. 
 
Similarly to the sentence-level retrieval, we use the HN-NCE contrastive formulation defined in Eq.(\ref{eq:hnnce}) for the sign retrieval (SignRet) loss. 
However, instead of the \emph{full sentence} video-text embedding pair $(\mathbf{V}, \mathbf{T})$,
we map \emph{individual} sign video-word embedding pairs $(\mathbf{v}, \mathbf{t})$ (see Fig.~\ref{fig:method}\textcolor{red}{c}). 

\noindent\textbf{Overall loss.} Our model is trained jointly using a weighted sum of the two retrieval terms:
\begin{align*}
    \mathcal{L} = \lambda_{\text{SentRet}}\mathcal{L}_{\text{SentRet}}
    + \lambda_{\text{SignRet}} \mathcal{L}_{\text{SignRet}}
\end{align*}
with $\mathcal{L}_{\text{SentRet}}, \mathcal{L}_{\text{SignRet}}$, i.e.\ two contrastive losses
for sentence and sign retrieval, respectively.
The training details including the batch size, learning rate, and other hyperparameters can be found in the supplementary materials.

\subsection{Sources of supervision}
\label{subsec:supervision}

Leveraging weak and noisy text labels for the CSLR and retrieval training constitutes 
one of the key contributions of this work. 
Next, we present our two sources of
text supervision, namely, sign-level pseudo-labels and sentence-level weakly-aligned subtitles.

\noindent\textbf{Sign-level pseudo-labels.}
We start with $(V, T)$ video-subtitle pairs that do not contain sign-level annotations.
In order to obtain sign-level supervision to train for CSLR, we perform sign-level pseudo-labelling. Specifically, we apply an ISLR model
in a sliding window fashion with a stride of 2 frames, and perform post-processing of sign predictions as an attempt to reduce noise. Our post-processing strategy consists of 3 steps: (i) we first combine confidence scores of synonym categories for the Top-5 predictions from the ISLR model (using the synonym list defined in~\cite{Momeni22}); (ii) we then filter out low confidence predictions (below a threshold value of $\theta = 0.6$); (iii) finally, we remove non-consecutive predictions -- since each sign spans several video frames, we expect repetitions from the ISLR model (we keep predictions with at least $m = 6$ repetitions). 

\cvprmodifs{In practice, for each subtitle, we define a sentence-level video (on average 3.4 seconds) by trimming the episode-level video ($\sim$1h duration)
using the subtitle timestamps.
The sentence-level video
is further broken down into frame-sign correspondences based on pseudo-label timestamps.
The sign-level loss is then only computed on frames associated to a pseudo-label after post-processing.
}

\noindent\textbf{Weakly-aligned subtitles.}
The source of our large-scale video-subtitle pairs is from sign language interpreted TV shows, where the timings of the accompanying subtitles correspond to the audio track, but not necessarily to signing \cite{Albanie21b}. 
For better sign-video-to-text alignments, we use automatic signing-aligned subtitles from~\cite{Bull21} (described in \cite{Albanie21b}) to train our models.
We restrict our training to subtitles spanning 1-20 seconds, resulting in 689K video-subtitle training pairs.

\subsection{Implementation details}
\label{subsec:implementation_details}

In the following, we detail each component of our model.

\noindent\textbf{Sign video encoder (\vencsign).} Similar to~\cite{Prajwal22a}, our sign video features are obtained by training a Video-Swin model~\cite{Liu2022VideoST}, 
on ISLR. 
The network ingests a short video clip (16 frames, $< 1$ second)
and outputs a single vector $\mathbf{\hat{v}} \in \mathbb{R}^d$ ($d=768$), followed by a classification head to recognise isolated signs.
\cvprmodifs{Specifically, we finetune the Video-Swin-Tiny architecture, pretrained on Kinetics-400~\cite{carreira2017quo}, using automatic annotations released in~\cite{Momeni22}.
These annotations provide individual sign labels along with timestamps of where they occur in the video.
}

Note that the annotations have been \textit{automatically} obtained with the help of subtitles (by exploiting cues such as mouthing),
and can thus be noisy. 
Once trained for ISLR, 
we freeze the parameters of this relatively expensive backbone,
and extract isolated sign video embeddings $\mathbf{\hat{v}}$ 
in a sliding window manner \cvprmodifs{with a stride of 2 frames}.
\cvprmodifs{Note we use RGB-based embeddings, instead of body keypoint estimates,
due to their more competitive performance
in the large-vocabulary setting, where sign differences are subtle and nuanced \cite{Albanie21b}.}

\noindent\textbf{Sentence video encoder (\vencsent).}
We adopt a Transformer encoder architecture, similar to BERT~\cite{devlin2018bert},
with \cvprmodifs{6 encoder layers, 8 attention heads and 768 hidden dimensionality.}
It ingests the sign sentence video as isolated sign video embeddings $\{\hat{\mathbf{v}}_f\}$ and outputs context-aware sign video embeddings $\{\mathbf{v}_f\}$. We obtain a single sentence video embedding by simply max-pooling over the temporal dimension, i.e.\ $\mathbf{V} = \text{MaxPool}_f(\{\mathbf{v}_f\}_{f=1}^F)$, with $F$ video features, and experimentally validate this choice. 
\cvprmodifs{We restrict our training to video clips shorter than 20 seconds.
} 
The parameters of the Transformer encoder  are learnt 
using the sentence retrieval loss $\mathcal{L}_{\text{SentRet}}$  between sentence text~$T$
and video $V$ embeddings, and the sign retrieval loss $ \mathcal{L}_{\text{SignRet}}$ between 
sign text embeddings $\mathbf{t}_w$ and the corresponding sign video embeddings $\mathbf{v}_f$,
as described in Sec.~\ref{subsec:objectives}.

\noindent\textbf{Text encoder (\tenc).} We use the encoder part of a \cvprmodifs{pre-trained T5~\cite{raffel2020t5} (\texttt{t5-large}), 
and keep its weights frozen. Note we do not use its decoder. The output text embeddings
have dimensionality 1024.}

\noindent\textbf{Projection heads.}
\cvprmodifs{We additionally learn projection layers, mainly to reduce the joint embedding dimensionality to 256 before contrastive loss computations.
	Specifically, we have a total of 4 projection heads: two for reducing the \textit{text} dimensionality ($1024 \rightarrow 256$) with a separate projection for sign categories and sentences, two for reducing the \textit{video} embedding dimensionality ($768 \rightarrow 256$) separately for sign and sentence video embeddings.
}

	\section{A New CSLR Evaluation Benchmark}
\label{sec:data}

In this section, we describe the new continuous sign annotations that we collected for evaluating CSLR. We first describe what the \testCSLR is, and then how it was annotated.

\noindent\textbf{\testCSLR.}  
The continuous annotations are provided for a subset of the \testManual partition of
BOBSL~\cite{Albanie21b}.  \testManual is a 31 hour subset of the BOBSL
test set where the BSL signing sequences have been manually  aligned temporally with
their corresponding English subtitle sentences.

The \testCSLR annotations consist of a time aligned sequence of sign `glosses'\footnote{We abuse the gloss terminology, despite our sign-level annotations \textit{not} being careful linguistic glosses, but rather free-form \cvprmodifs{sign-level} translations.}, where each sign is annotated with its temporal interval, the type 
of the sign, and its word equivalent if that exists. 
\cvprmodifs{In addition to lexical signs (i.e.\ signs that have an English word equivalent), 
a wide range of sign types such as fingerspelling, pointing, depicting or no-signing are annotated. 
These are marked with special characters such as *FS and *P for fingerspelling and pointing, respectively.
}

\begin{figure*}
    \centering
    \includegraphics[width=.98\linewidth]{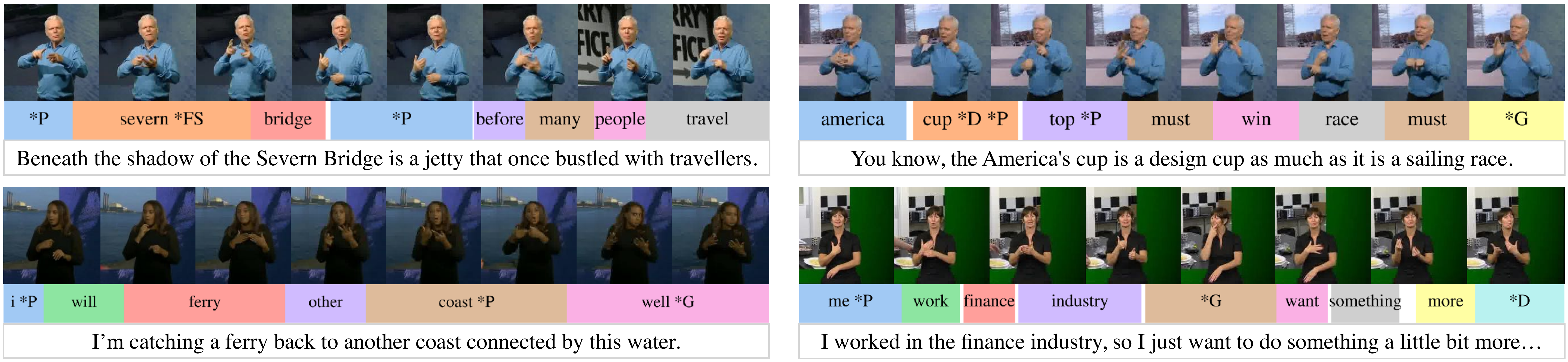}
    \vspace{-0.3cm}
     \caption{
        \cvprmodifs{\textbf{Annotation examples from the \testCSLR dataset:} As well as assigning the English word(s) corresponding to a sign (i.e.\ `gloss'), the annotators indicate the type of sign when appropriate. For example, `*P' for pointing, `*FS' for fingerspelling, `*G' for gesture sign.
        }
    }
    \vspace{-0.5cm}
    \label{fig:examples-sign-annot}
\end{figure*}

Note, that there exists no universally accepted writing system for sign languages today~\cite{filhol2020elicitation}, though 
attempts have been made with descriptive languages such as
HamNoSys \cite{hanke2004hamnosys} and SignWriting \cite{sutton1990signwriting}.
Also, careful glossing that is linguistically consistent (e.g., enumerating
each sign variant~\cite{bslcorpus17})
is a tedious process which hinders scaling up.
For these reasons, we make a compromise when annotating for CSLR and use English words for the glosses,
assigning `any' reasonable English word for a sign segment, but prioritising words in the surrounding
subtitle. For example, if the `natural' English word for the sign is `laugh' but a 
synonym word such as `giggle' is in the subtitle, then the gloss would be `giggle' (with `laugh' also provided as a 
more general translation).
However, one should keep in mind that associating words to signs is a lossy
and error-prone process in any case. 

Fig.~\ref{fig:examples-sign-annot} shows example ground truth gloss annotations for \testCSLR.
In total, we curate these continuous labels for \cvprmodifs{6 hours of video,
comprising 32.4K individual signs from a vocabulary of
approximately 5.1K glosses.}
The \testCSLR annotations evenly cover all 35 episodes in \testManual.
Additional statistics for the dataset are given in Tab.~\ref{tab:datasets}. We note that we also annotate a small subset of the BOBSL training and validation subtitle-aligned splits. All annotations will be publicly released. 

\noindent\textbf{Dataset annotation.}
The annotation procedure uses a web based annotation tool that is built from the VIA video annotation 
software~\cite{Dutta19a}. Annotators are provided with a video sequence with 10 time aligned subtitle sentences.
Annotators enter free-form text for each sign token, taking into account
the context by watching the full video around a given subtitle. The subtitle is also displayed on the video and the annotators are encouraged to prioritise assigning words that appear in the corresponding
subtitle.  
The annotators additionally assign sign types when appropriate (see Fig.~\ref{fig:examples-sign-annot}) and temporally align each gloss to the duration of the sign.

We facilitate faster annotation iterations by incorporating
several strategies.
We adopt a semi-automatic labeling technique where we initialise the sign boundaries by using an automatic
sign segmentation method 
\cite{Renz21a}.
Annotators are thus given an initial set of sign intervals, and are instructed to
refine these sign boundaries (or add/remove sign intervals) if necessary.
We further show a dropdown menu for each sign 
and prioritise at the top of the list words from the corresponding subtitle.
Two native BSL users worked on this task for \cvprmodifs{over one year}. 
Further details on the annotation procedure are provided in the supplementary material. 

	\section{Experiments}
\label{sec:experiments}

In this section, 
we first present evaluation protocols
used in our experiments (Sec.~\ref{subsec:evaluation}) and describe baselines (Sec.~\ref{subsec:baselines}).
Next, we provide ablations to assess the contribution
of important components (Sec.~\ref{subsec:ablations}).
We then report CSLR and retrieval performance, comparing to the state of the art (Sec.~\ref{subsec:sota}), and 
illustrate qualitative results (Sec.~\ref{subsec:qualitative}).

\subsection{Data and evaluation protocol}
\label{subsec:evaluation}

\textbf{BOBSL}~\cite{Albanie21b} consists of about 1500 hours of video data
accompanied with approximately-aligned subtitles. A 200-hour subset
is reserved for testing. 
We reuse existing manually-aligned validation and test sets (\textbf{\valManual~\cite{Albanie21b}}, \textbf{\testManual~\cite{Albanie21b}})
for our sentence retrieval evaluation (20,870 and 1,973 aligned sentences respectively). 
We perform our retrieval ablations on the validation set, and report the final model on both evaluation sets.
For CSLR evaluation, we use our
manually annotated test set (\textbf{\testCSLR})
as described in Sec.~\ref{sec:data}, which corresponds to 4950 
unseen test subtitles. 

For retrieval evaluation, we report both T2V and V2T performances using standard retrieval metrics, namely \textbf{recall at rank k (R@k)} for k $\in \{1, 5\}$. 
For CSLR evaluation, given a video sequence defined by \testCSLR, we compute our predicted gloss sequence (after post-processing the raw per-frame outputs with the optimal $\theta$ and $m$ heuristics -- see Sec.~\ref{subsec:supervision} for post-processing strategy) and
compare against its corresponding ground-truth gloss sequence
using several metrics.
We note that we filter out sign types and signs that are not associated to lexical words from the ground-truth sequence.
In addition, if several annotation words are associated to one sign (e.g.\ `giggle' and `laugh'), predictions are considered to be correct if one of these words is predicted correctly.

As in other CSLR benchmarks~\cite{Koller15cslr,Zhou2021ImprovingSL},
we report word error rate (\textbf{WER}) as our main performance measure.
We also monitor \textbf{mIoU} (mean intersection over union) \cvprmodifs{between predicted and ground truth sequences' words, without considering any temporal aspect.}
In all metrics,
similar to~\cite{Momeni22},
we do not penalise output words if they are synonyms.

\cvprmodifs{
In order to assess the model's ability to correctly predict sign segments at the right temporal location, 
we also report the F1 score.
We define a segment to be correctly predicted if (i) the predicted sign matches, up to synonyms,
the ground-truth gloss, and (ii) the IoU between the predicted segment's boundaries
and the ground truth gloss segment's boundaries is higher than a given threshold.
We compute the F1 score as the harmonic mean of precision and recall, 
based on this definition of correct segment detection.
We report the F1 score at different thresholds values, namely, $\mathbf{\text{F}1@\{0.1, 0.25, 0.5\}}$
}.

\subsection{Baselines}
\label{subsec:baselines}
\noindent\textbf{Subtitle-based automatic annotations CSLR baseline.} 
The first baseline for CSLR is obtained using spotting methods that search for signs corresponding to words in the spoken language subtitles. 
The initial set of sparse sign annotations along with timestamps was released with the BOBSL dataset~\cite{Albanie21b},
using sign spotting methods from~\cite{Albanie20,Momeni20b,Varol21}, followed by denser spottings in~\cite{Momeni22}. 
We evaluate these existing sequences of spottings, in particular the ones corresponding to our \testCSLR subtitles.
In practice, we filter these automatic annotations using the same sets of thresholds as in~\cite{Albanie21b,Momeni22}, respectively (see supplementary material for details).
Note that these spottings make use of the weakly-aligned subtitles (and cannot go beyond words in the subtitles), and therefore cannot be used as a true CSLR method. They are also point annotations, without precise temporal extent, thus we omit F1 scores.

\noindent\textbf{ISLR baselines for CSLR.} 
This set of baselines uses ISLR models in a sliding window fashion to obtain continuous frame-level predictions. We aggregate the sliding window outputs by performing the post-processing strategy described in Sec.~\ref{subsec:supervision} (for optimal $\theta$ and $m$ parameters, tuned on unseen manually annotated CSLR sequences from the validation set). 
We use 
the I3D~\cite{carreira2017quo} and Video-Swin~\cite{Liu2022VideoST} models 
trained from spotting annotations of~\cite{Albanie21b,Momeni22}.
We build on prior works for these baselines: we use the I3D weights released in~\cite{Albanie21b} and train a 
Video-Swin-Tiny
model in a similar fashion as in~\cite{Prajwal22a}. 

\noindent\cvprmodifs{\textbf{InfoNCE retrieval baseline.} Our baseline for sentence
retrieval is the standard contrastive training~\cite{infonce} employed
by many strong vision-language models~\cite{clip2021,BLIP2022}. 
We train this vanilla model, without the CSLR objective, on the automatically-aligned subtitles from~\cite{Bull21}.
Sentence embeddings, obtained by feeding subtitle text into \tenc, are compared against a learnable \texttt{cls} token on the video side which pools the video embeddings as done in~\cite{dosovitskiy2021an,clip2021}.}

\subsection{Ablation study}
\label{subsec:ablations}
\begin{table}
    \centering
    \setlength{\tabcolsep}{8pt}
    \resizebox{1\linewidth}{!}{
    \begin{tabular}{c|c|ccccc}
        \toprule 
        SentRet & Sign-level loss & WER $\downarrow$ & mIOU $\uparrow$ & \multicolumn{3}{c}{F1@$\{0.1, 0.25, 0.5\} \uparrow$} \\
        \midrule
        \rowcolor{goodcolor}\multicolumn{2}{l}{\textsc{ISLR Baseline}} & 71.7 & 30.1 & 40.0 & 37.9 & 27.2 \\
        \midrule
        \xmark & CTC & 75.0 & 27.8 & - & - & -  \\
        \xmark & CE & 72.9 & 28.6 & 37.9 & 37.3 & 29.1 \\
        \xmark & SignRet & 71.2 & 30.4 & 40.7 & 39.9 & 31.0 \\
        \midrule
        \rowcolor{aliceblue} \cmark & CE & 71.0 & 30.4 & 39.7 & 38.9 & 30.9  \\
        \rowcolor{aliceblue} \cmark & SignRet & \textbf{65.5} & \textbf{35.5} & \textbf{47.1} & \textbf{46.0} & \textbf{37.1} \\
        \bottomrule
    \end{tabular}
    }
	\caption{\textbf{CSLR ablations on \testCSLR 
    :}
	Only using CTC, cross entropy (CE), or SignRet does not perform well, remaining below or comparable to the ISLR baseline.
	We observe best results when incorporating joint sentence retrieval training.
	}
	\label{tab:cslr_ablations}
	\vspace{-0.4cm}
\end{table}
\noindent\textbf{CSLR components.}
In Tab.~\ref{tab:cslr_ablations}, we experiment with the choice 
of training objectives for CSLR performance.
\cvprmodifs{In particular, we train with the standard CTC or cross-entropy (CE) losses,
	as well as our sign retrieval (SignRet) loss alone. 
Without an additional sentence retrieval loss, i.e.~if we only optimise 
for a sign-level objective,
we observe that the performance is worse than the strong ISLR baseline
(i.e.\ with Video-Swin)
for CTC and CE, and comparable for SignRet.
In the final two rows, we observe clear gains by combining the SentRet loss with either
(i)~our SignRet loss or (ii)~the standard CE loss. 
While the joint training with the CE loss brings a performance boost, from 72.9 to 71.0 WER, it does not significantly surpass the competitive ISLR baseline. Our model \modelname, which jointly trains sentence and sign retrieval, brings a major improvement by reducing the WER by 5.7 points, from 71.2 to 65.5.
}

\begin{table}
    \centering
    \setlength{\tabcolsep}{4pt}
    \resizebox{1\linewidth}{!}{
        \begin{tabular}{c|c|c|ccc|ccc} 
            \toprule 
            & SentRet & Sign-level & \multicolumn{3}{c|}{T2V} & \multicolumn{3}{c}{V2T} \\ 
            Pool. & loss & loss & R@1 $\uparrow$ & R@5 $\uparrow$ & R@10 $\uparrow$ & R@1 $\uparrow$ & R@5 $\uparrow$ & R@10 $\uparrow$ \\ 
            \midrule
            \rowcolor{goodcolor} \texttt{cls}  & InfoNCE & \xmark & 38.9 & 62.1 & 69.1 & 39.2 & 61.0 & 68.1 \\
            \texttt{cls} & HN-NCE & \xmark & 48.9 & 68.3 & 73.9 & 46.5 & 67.2 & 72.8 \\
            \midrule
            \texttt{max} & InfoNCE & \xmark & 43.4 & 64.9 & 71.0 & 42.7 & 64.8 & 70.7 \\
            \texttt{max} & HN-NCE & \xmark & 50.5 & 69.5 & 75.1 & 49.7 & \textbf{69.7} & \textbf{74.7} \\
            \midrule
            \rowcolor{aliceblue}\texttt{max} & HN-NCE & CE & 50.0 & 69.1 & 74.4 & 48.7 & 68.7 & 74.3 \\
            \rowcolor{aliceblue}\texttt{max} & HN-NCE & SignRet & \textbf{51.7} & \textbf{69.9} & \textbf{75.4} & \textbf{50.2} & 69.1 & \textbf{74.7} \\
        \bottomrule
        \end{tabular}
    }
	\caption{\textbf{Retrieval ablations on \valManual:}
	We experiment with the choice of the contrastive sentence retrieval (SentRet) loss (standard InfoNCE vs.\ HN-NCE), the visual encoder pooling (\texttt{cls} vs \texttt{max}),  the addition and choice of sign-level losses (cross entropy CE vs.\ contrastive sign retrieval SignRet). The last two rows correspond to the joint models evaluated for CSLR in Tab.~\ref{tab:cslr_ablations}.
	}
	\label{tab:retrieval_ablations}
\end{table}

\noindent\textbf{Retrieval components.}
In Tab.~\ref{tab:retrieval_ablations}, we compare design choices for the retrieval task: 
(i)~choice of the contrastive loss function -- InfoNCE vs.~HN-NCE; 
(ii) choice of pooling the temporal features -- using a \texttt{cls} token~\cite{devlin2018bert,dosovitskiy2021an} vs.~\texttt{max}-pooling. 
We observe a clear boost in all metrics by using HN-NCE with our weakly-aligned data, which gives more weight to the hard-negatives when computing the contrastive loss: there is a minimum improvement of +7 R@1 for T2V comparisons. We further observe that \texttt{max}-pooling the visual Transformer encoder outputs, instead of using a learnable \texttt{cls} token, consistently gives better results.
The joint training of retrieval and CSLR, with SignRet, also improves retrieval performance: R@1 for T2V increases from 50.5 to 51.7.
More importantly, joint training enables a single, strong model which can perform both tasks.

\begin{figure*}
    \centering
    \includegraphics[width=\linewidth]{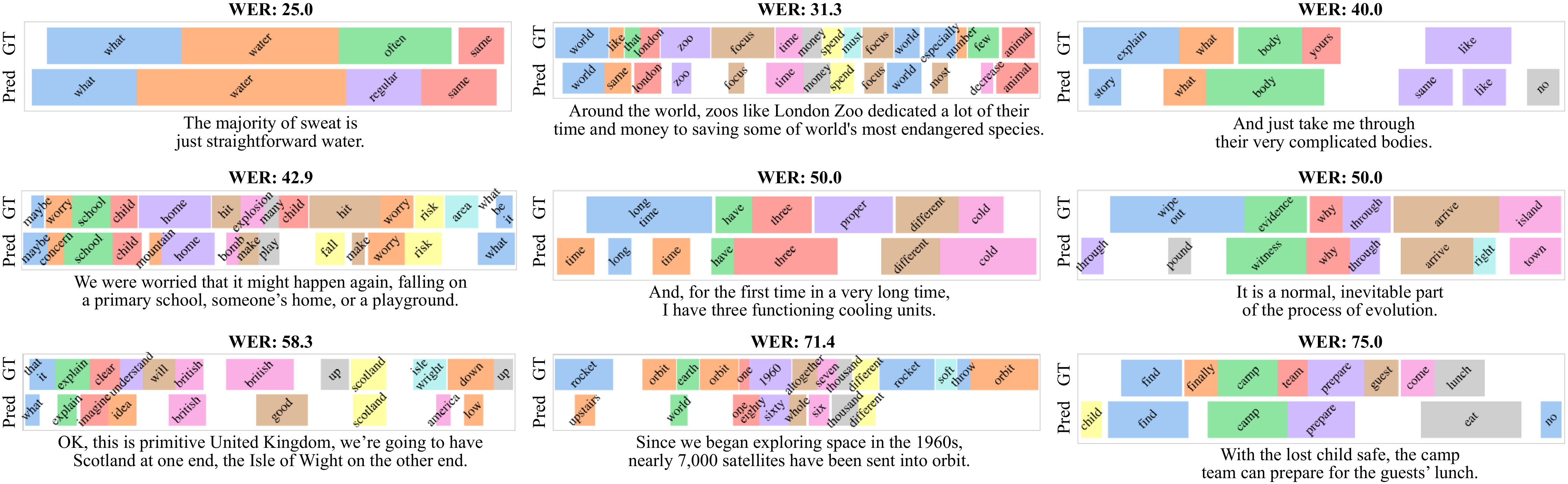}
    \vspace{-0.7cm}
    \caption{
        \textbf{Qualitative CSLR results:} We compare our model's predictions (Pred) against the ground truth (GT), providing examples from several error ranges (sorted by WER). The subtitles displayed below each example are not used by the model. While we observe that our model correctly predicts a large portion of signs, handling both English synonyms as well as sign language polysemy (two visually similar signs with different meanings) makes the CSLR task challenging. 
        \cvprmodifs{Synonyms are depicted with the same color coding, e.g.\ `earth' and `world' in 3rd row, middle.}
    }
    \label{fig:qualitative}
    \vspace{-0.5cm}
\end{figure*}

\begin{table}
    \centering
    \begin{subtable}{1.\linewidth}
        \centering
        \setlength{\tabcolsep}{8pt}
        \resizebox{\linewidth}{!}{
            \begin{tabular}{lccccc}
            \toprule
            \textsc{\bf CSLR} & \multicolumn{5}{c}{\testCSLR} \\
            Model & WER $\downarrow$ & mIOU $\uparrow$  & \multicolumn{3}{c}{F1@$\{0.1, 0.25, 0.5\} \uparrow$} \\
            \midrule
            \quad Subtitle-based spotting~\cite{Albanie21b} & 93.8 & \textcolor{white}{0}7.1 & - & - & - \\ 
            \quad Subtitle-based spotting~\cite{Momeni22} & 85.5 & 18.1  & - & - & - \\
             \quad ISLR I3D-2K~\cite{Albanie21b} & 453.5 & \textcolor{white}{0}8.6 & 10.6 & \textcolor{white}{0}8.5 & \textcolor{white}{0}5.4 \\
             \arrayrulecolor{lightgray}
             \cmidrule(l{1.5em}r{0em}){1-6}
             \arrayrulecolor{black}
            \quad ISLR I3D-2K~\cite{Albanie21b} $\dagger$ & 82.5 & 18.1 & 25.7 & 23.8 & 16.0 \\
            \quad ISLR I3D-8K~\cite{Momeni22} $\dagger$ & 74.7 & 27.0 & 36.7 & 35.4 & 26.7 \\
            \rowcolor{goodcolor} \quad ISLR Swin-8K~\cite{Prajwal22a} $\dagger$ & 71.7 & 30.1 & 40.0 & 37.9 & 27.2 \\
            \midrule
            \rowcolor{aliceblue} \quad \modelname \textsc{(Ours)} $\dagger$ & \textbf{65.5} & \textbf{35.5}  & \textbf{47.1} & \textbf{46.0} & \textbf{37.1} \\
            \bottomrule
        \end{tabular}
        }
        \caption*{} 
        \vspace{-6pt} 
    \end{subtable}
    
    \begin{subtable}{1.\linewidth}
        \centering
        \setlength{\tabcolsep}{8pt}
        \resizebox{\linewidth}{!}{
            \begin{tabular}{lcc|cccc}
            \toprule
            \textsc{\bf Retrieval} & \multicolumn{2}{c|}{\valManual~(2K)} & \multicolumn{4}{c}{\testManual (20K)} \\
            & T2V & V2T & \multicolumn{2}{c}{T2V} & \multicolumn{2}{c}{V2T} \\
            Model & R@1 $\uparrow$ & R@1 $\uparrow$ & R@1 $\uparrow$ & R@5 $\uparrow$ & R@1 $\uparrow$ & R@5 $\uparrow$ \\
            \midrule
            \rowcolor{goodcolor} \quad InfoNCE & 38.9 & 39.2 & 19.5 & 35.1 & 18.9 & 33.8 \\
            \midrule
            \rowcolor{aliceblue} \quad \modelname \textsc{(Ours)} & \textbf{51.7} & \textbf{50.2} & \textbf{29.4} & \textbf{45.2} & \textbf{28.1} & \textbf{44.9} \\
            \bottomrule
        \end{tabular}
        }
        \caption*{} 
        \vspace{-20pt}
    \end{subtable}
    \caption{\textbf{Comparison to the state of the art:} 
	 Our joint model significantly outperforms both CSLR (top) and retrieval (bottom) baselines.
	 For CSLR, note that the automatic
	 spotting annotations~\cite{Albanie21b,Momeni22} have \textit{access to the subtitles} at inference (unlike our fully automated approach).
	 We also compare to raw ISLR outputs from sign classification models from~\cite{Albanie21b,Momeni22,Prajwal22a} with various backbones (I3D or Swin) and with various vocabularies (2K or 8K categories). 
	 Our optimal filtering and post-processing strategy at inference is denoted with $\dagger$ (see.~\ref{subsec:supervision}). We note that for the ISLR I3D-2K baseline without~$\dagger$, we still remove consecutive repetitions.
	} 
    \label{tab:cslr_retrieval}
    \vspace{-0.4cm}
\end{table}

\subsection{Comparison to the state of the art}
\label{subsec:sota}

We compare to the current state-of-the-art approaches, both for large-vocabulary CSLR and sentence retrieval, in Tab.~\ref{tab:cslr_retrieval}.

\noindent \textbf{CSLR performance.}
First, in terms of the baselines, it can be seen from  Tab.~\ref{tab:cslr_retrieval}, that our post-processing strategy significantly strengthens the original ISLR I3D-2K~\cite{Albanie21b} performance by removing significant noise (with/without~$\dagger$) --  we reduce the WER by more than a factor of 5 (453.5 vs 82.5). Also,  our post-processing strategy combined with the 8K vocabulary ISLR models, delivers models of higher performance (by more than 10 WER) than all subtitle-based spottings methods, even though the ISLR models do not have access to the subtitles.
Second, our joint model, \modelname, outperforms all CSLR baselines by a significant margin on all metrics.
Indeed, \modelname surpasses the best subtitle spotting method by 20 WER and the strongest ISLR baseline by 6.2 WER. Please refer to the supplementary material for a breakdown of performance based on different sign types. 

\noindent \textbf{Retrieval performance.}
As Tab.~\ref{tab:cslr_retrieval} shows, our joint \modelname model outperforms a standard InfoNCE~\cite{infonce} baseline for retrieval on all reported metrics, with  gains in R@1 for both T2V and V2T of almost 10 points.
On the more challenging \testManual gallery of 20k video-subtitle pairs,
our \modelname model achieves a Top-5 accuracy of 45.2\% for T2V retrieval. We observe that for cases where the target sentence is not the Top-1, the top-retrieved results  usually exhibit semantic similarities with the correct sentence, with multiple common words
(see the qualitative examples in the supp.~mat.).

\subsection{Qualitative analysis}

\label{subsec:qualitative}
\cvprmodifs{
In Fig.~\ref{fig:qualitative}, we show several qualitative examples of our CSLR predictions (Pred rows)
against the corresponding ground truth (GT rows) on \testCSLR.
Note that we display the corresponding ground truth subtitles below each example
to give context to the reader, but they are not given as input
to the model.
We illustrate examples from several error ranges, sorting them by WER per sample (reported at the top).
These timelines show that our model is able to predict a large proportion of the annotations, 
in the correct order, with approximate sign segmentation.
For instance, even though the bottom-right example in Fig.~\ref{fig:qualitative} has a high error rate of 75 WER, our predictions  correctly identify 4 out of the 8  ground truth words, and catch the meaning of the sentence.
}

\cvprmodifs{
However, we also observe several challenges: 
(i)~our model has difficulty predicting several words for a single sign, as the 8K training vocabulary of pseudo-labels primarily comprises of individual words
(e.g.\ the phrase `long time' is associated to a single sign in the 2nd row - middle - but our model predicts two separate signs `long' and `time', leading to an extra insertion)
(ii)~our model performance is sensitive to the synonyms list, which must be carefully constructed to not unfairly penalise predictions
(e.g.\ in the top left example, `regular' is counted as a substitution since `often' is not present in `regular''s synonyms list)
(iii)~our model still struggles with visually similar signs in BSL which correspond to different English words
(e.g.\ `upstairs' and `rocket' signs are visually similar, both signed by pointing upwards, 3rd row - middle);
(iv)~finally, our model is more likely to fail in recognising names of places and people 
as these are often fingerspelled in BSL and may therefore not be in our 8K sign vocabulary of pseudo-labels
(e.g.\ `isle wright' is predicted as `america' in the bottom left ).} Future directions include addressing such limitations.

	\section{Conclusion}
\label{sec:conclusions}
In this work, we demonstrate that jointly training for CSLR and sign language retrieval is mutually beneficial. We collect a large-vocabulary CSLR benchmark, consisting of 6 hours of continuous sign-level annotations. By leveraging weak supervision, we train a single model which outperforms strong baselines on both our new CSLR benchmark and existing retrieval benchmarks. While our approach shows substantial improvements, future work includes increasing the vocabulary size beyond 8K and modeling non-lexical signing classes such as pointing and gesture-based signs.

\noindent\textbf{Societal impact.}
The two sign language understanding tasks we address can have 
positive implications by bridging the gap between spoken and sign languages. These tasks can enable more seamless communication, content creation and consumption by breaking down the language barriers that are prevalent today. 
At the same time,
the ability to automatically search a large volume of signing videos can lead to risks such as surveillance of signers.
We believe that the positives outweigh the negatives.

\medskip
{ \small
\noindent\textbf{Acknowledgements.}
This work was granted access to the HPC resources of IDRIS under the 
allocation {2023-AD011013569} made by GENCI.
The authors would like to acknowledge
the ANR project CorVis ANR-21-CE23-0003-01.
\par
}
	
	{
		\small
		\bibliographystyle{ieeenat_fullname}
		\bibliography{shortstrings,vgg_local,references}
	}
	\clearpage
        \bigskip
        {\noindent \large \bf {APPENDIX}}\\
	\appendix

\renewcommand{\thefigure}{A.\arabic{figure}}
\setcounter{figure}{0} 
\renewcommand{\thetable}{A.\arabic{table}}
\setcounter{table}{0}

\startcontents[sections]
{
	\hypersetup{linkcolor=black}
	\printcontents[sections]{l}{1}{}
}

\section{Implementation details}\label{section:app-imp-details}

We provide additional implementation details to what is described in
\if\sepappendix1{Sec.~3.4}
\else{Sec.~\ref{sec:method}}
\fi
of the main paper.

\begin{table*}[ht]
    \centering
    \setlength{\tabcolsep}{5pt}
    \resizebox{1\linewidth}{!}{
    \begin{tabular}{p{.1\linewidth}|p{0.47\linewidth}|c|c}
    \toprule
          Notation & Description & Input & Output \\
          \midrule
         \vencsign &  Video-Swin-Tiny (across $T$ clips) & $T\times16\times224\times224\times3$ & $T\times768$ \\
         \vencsent & Transformer Encoder with 6 encoder layers, 8 attention heads and 768 hidden dimensions & $B \times F \times 768$ & $B \times F \times 768$\\
         \tenc & Pre-trained T5~\cite{raffel2020t5} ("t5-large") encoder & $B\times W$ & $B \times 1024$ (sentence-level) \\
         & & & $B \times W \times 1024$ (word-level)\\
         \midrule
         FC & Two linear projections from 768-d space to 256-d space & $B \times F \times 768\text{\textcolor{white}{0}}$ & $B \times F \times 256$ \\
         & Two linear projections from 1024-d space to 256-d space & $B \times F \times 1024$ & $B \times F \times 256$ \\
    \bottomrule
    \end{tabular}
    }
        \vspace{-0.3cm}
	\caption{
	\textbf{Model summary:} $F$ is the maximum number of frames for all video sequences in a batch of size $B$. Zero-padding is applied to inputs. $W$ is the maximum number of tokens obtained after tokenisation of $B$ sentences. Zero-padding is also applied here.
	}
    \label{tab:app:model_details}
\end{table*}

\noindent \textbf{\modelname training details.} 
We perform the contrastive training with batch size 512 and learning rate $\text{lr} = 5 \times 10^{-5}$ using Adam~\cite{kingma2014adam} optimizer. The HN-NCE loss uses temperature $\tau = 0.07$, $\alpha = 1.0$ and $\beta_{\text{SignRet}} = 0.5$, $\beta_{\text{SentRet}} = 1.0$.
We use $\lambda_{\text{SignRet}} = 0.0075$, $\lambda_{\text{SentRet}} = 0.90$ to weigh the different terms in our joint loss. 
Training takes less than 20 hours on 4 NVIDIA 32 GB V100 GPUs.
We choose the model checkpoint that performs best for T2V R@1 on a manually aligned subset of validation subtitles. 

\noindent \textbf{Data augmentation} is performed on both the video and text side. 
On the text side, random words from the subtitles are dropped using two probabilities.
First, we choose with probability $p_{\text{sub}, \text{drop}} = 0.8$ whether words will be dropped in a given subtitle.
Second, in subtitles selected by the previous step, we drop each word with probability $p_{\text{word}, \text{drop}} = 0.4$. 
On the video side, we perform dropping in a similar fashion to dropping words in subtitles, 
and drop features in sequences with two probabilities: $p_{\text{seq}, \text{drop}} = 0.8$ and $p_{\text{frame}, \text{drop}} = 0.5$.
We also temporally shift subtitles with a random offset, which is uniformly sampled in the range $\pm 0.5$ seconds.

\noindent \textbf{Model summary} is displayed in Tab.~\ref{tab:app:model_details} (please also refer to
\if\sepappendix1{Fig.~2}
\else{Fig.~\ref{fig:method}}
\fi
of the main paper).

\noindent \textbf{Video-Swin-Tiny training hyper-parameters.} We train the video backbone to perform isolated sign language recognition on short video clips of $16$ frames, starting from Kinetics-400 pre-trained weights\footnote{\url{https://github.com/SwinTransformer/Video-Swin-Transformer}}. The input videos are of resolution $224\times224$, with the pixel values normalized by the ImageNet mean and standard deviation as also done in~\cite{Liu2022VideoST}. We only perform two data augmentations: (i) random horizontal flipping, (ii) random color jitter by scaling the original pixel values by a random factor between $[0.8, 1.2]$. We start with the Kinetics~\cite{carreira2017quo} pretrained weights and train for ISLR with a batch size of $256$ distributed across $4$ NVIDIA A40 GPUs over a period of 3 days. We use the Adam optimizer with the default parameters, except for the learning rate. The learning rate is linearly warmed up to a maximum of $1\times10^{-3}$ over $8000$ iterations, followed by a cosine decay schedule. We stop training when the model's validation accuracy (Top-1 and Top-5) does not improve for $2$ epochs. 

\section{\testCSLR}\label{section:app-cslr-test}
In this section, we provide further details on the annotation collection procedure for \testCSLR (\ref{appendix:annotation_procedure}) and dataset statistics (\ref{appendix:stats}).

\subsection{Annotation tool}\label{appendix:annotation_procedure}

\begin{figure*}[ht]
    \centering
    \begin{minipage}[c]{.6\linewidth}
        \centering
        \includegraphics[width=\linewidth]{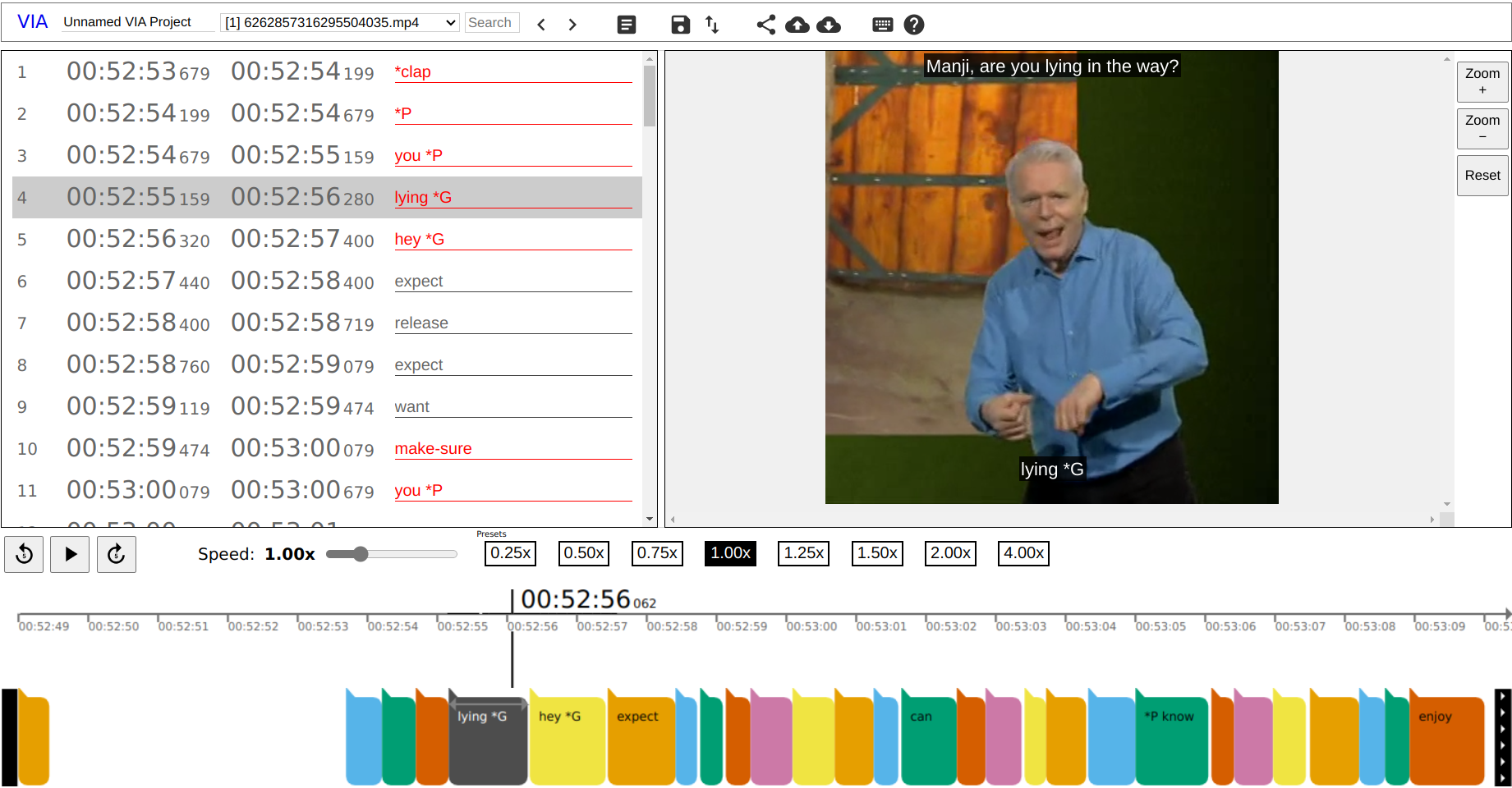}
        \subcaption{Final sign annotations}
    \end{minipage}
    \begin{minipage}[c]{.39\linewidth}
        \centering
        \includegraphics[width=\linewidth]{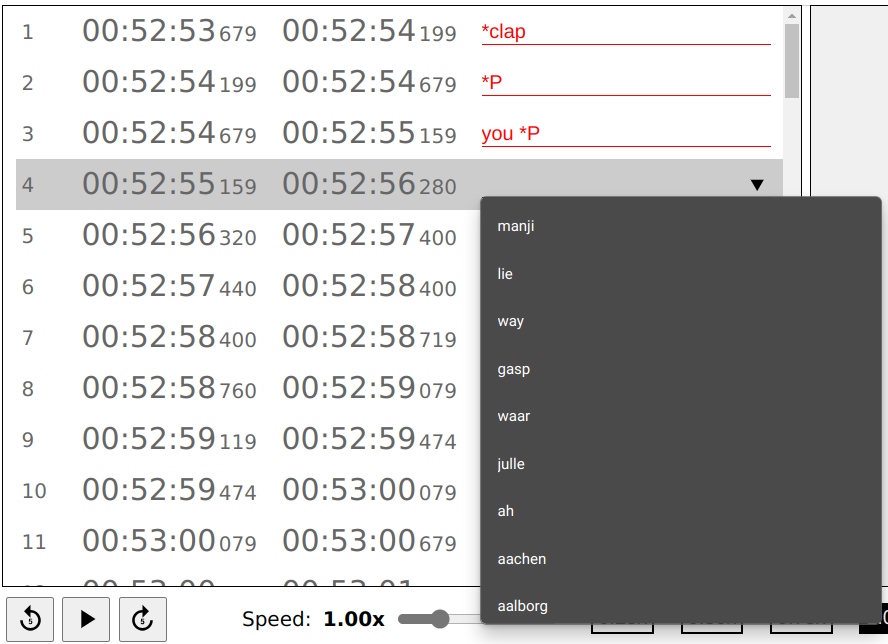}
        \subcaption{Zoom on dropdown menu before any entry}
    \end{minipage}
    \vspace{-0.3cm}
     \caption{
        \textbf{Annotation tool:} We show two screenshots of the web based annotation tool used for collecting \testCSLR. \textbf{(a)} The annotators use (i) the right-hand side of the tool to view the signing video and aligned English subtitle, (ii) the colored timeline at the bottom of the tool to adjust/add/remove sign intervals, (iii) the left-hand side to write free-form text to assign  English word(s) to sign intervals, as well as the sign type (for example, *P for pointing) when appropriate \cvprmodifs{and (iv) a dropdown menu for each sign with the vocabulary spanned by subtitle words in BOBSL.
        We note that on the bottom colored timeline, some sign intervals appear as if unlabelled simply because there is not enough space to fit the annotation text (however these are shown on the left hand side of the tool). 
        The segments are first initialised from automatic sign segmentation~\cite{Renz21b} and are refined by annotators. 
        \textbf{(b)} When no annotation is entered yet, subtitle words are displayed at the top of the drop down menu.} 
    }
    \label{fig:annot-tool}
\end{figure*}

In Fig.~\ref{fig:annot-tool}, we show two screenshots of the web-based annotation tool used for collecting \testCSLR. It is based on the VIA video annotation software~\cite{Dutta19a,Renz21a,Renz21b}. Specifically, we start from VIA-SLA\footnote{\url{https://github.com/RenzKa/VIA_sign-language-annotation}} and extend it for our purposes (by adding the signing-aligned subtitle, adjusting the dropdown vocabulary, and removing the Top-1 sign class prediction from a sign recognition model).

Annotators are provided with a video sequence with 20 signing-aligned subtitle sentences spanning roughly 2-3 minutes (this takes about 1 hour to annotate densely).
The subtitles are displayed at the top of the video. Annotators are given an initial set of sign intervals (shown at the bottom of the tool with a colored timeline) from an automatic
sign segmentation method~\cite{Renz21b}, but are instructed to refine these sign boundaries (or remove/add sign intervals) if necessary. 
For each sign interval, the annotators enter free-form text which best corresponds to the meaning of the sign, by taking into account context of the surrounding video. 
The annotators are encouraged to prioritise assigning words that appear in the corresponding subtitle. 
For example, in Fig.~\ref{fig:annot-tool}a, the signing-aligned subtitle is `Manji, are you lying in the way?' and the highlighted gray interval is annotated with the subtitle word `lying'.
We further show a drop down menu for each sign (with the full English vocabulary of the subtitle words in the BOBSL dataset, together with a search functionality),
with the subtitle words listed at the top of the list,
to enable faster annotation (see Fig.~\ref{fig:annot-tool}b). 
Besides assigning the English word(s) corresponding to a sign, the annotators indicate the \textit{type} of sign when appropriate. 
For example, as shown on the left of Fig.~\ref{fig:annot-tool}a, the third sign is annotated as `you *P' meaning the sign for `you' is a \textbf{p}ointing sign, and the fifth sign is annotated as `hey *G' meaning the sign for `hey' is a \textbf{g}esture sign. 
The full list of sign types annotated is given in the following section.

\subsection{Dataset statistics}\label{appendix:stats}

\noindent \textbf{BOBSL-CSLR.} In Tab.~\ref{tab:app:cslr-stats}, we show statistics for our collected \testCSLR data. We also collect annotations in a similar way for a small proportion of the train and validation splits.

\begin{table*}
    \centering
    \setlength{\tabcolsep}{5pt}
    \begin{tabular}{lccccccc}
            \toprule 
           & \#sentences& \#hours &  \#subtitle & subtitle & \#glosses& gloss & \#sign type \\
           &   &  & \textcolor{white}{\#s}words & \textcolor{white}{s}vocab. &  & vocab. & \textcolor{white}{\#aa .}annot.  \\
            \midrule
            all & 7134 & 8.9 & 81.3K & 9.0K & 47.7K & 7.1K & 17.1K \\
            \midrule
            train & 1463 & 1.7 & 15.0K 
            & 3.1K & \textcolor{white}{0}8.7K & 2.6K & \textcolor{white}{0}3.4K \\
            val & 1153 & 1.3 & 13.4K
            & 2.7K & \textcolor{white}{0}6.6K & 2.1K & \textcolor{white}{0}2.0K \\
            test & 4518 & 6.0 & 52.9K
            & 6.9K & 32.4K & 5.1K & 11.7K \\
        \bottomrule
        \end{tabular}   
            \vspace{-0.3cm}
	    \caption{\textbf{BOBSL-CSLR:} We show statistics for our collected \testCSLR data (number of subtitle sentences, duration in hours, number and vocabulary of subtitle words, number and vocabulary of annotated glosses, number of annotated sign types). We also collect annotations in a similar way for a small proportion of the train and validation splits.
	}
	\label{tab:app:cslr-stats}
\end{table*}

\begin{table*}
    \setlength{\tabcolsep}{8pt}
    \centering
	\begin{tabular}{llcr}
        \toprule 
        Type of partially-lexical & Description & Short form &  \#occurences \\
        or non-lexical signs & & & in \testCSLR\\  
        \midrule
        Name sign & countries or people that  & *N & \textcolor{white}{00}75 \\ 
        & have a well-known BSL sign & & \\          
        \midrule
        Facial expression sign & meaningful facial expression & *FE & \textcolor{white}{0}252 \\ 
        \midrule 
        Slang sign & sign corresponding to slang word & *S & \textcolor{white}{0}258 \\ 
        \midrule 
        Timeline sign & signs which relate to time (present, past, & *T & \textcolor{white}{0}324 \\ 
        & future, calendar units, action continuity) & & \\
        \midrule 
        Unknown sign & unrecognisable signs & *U & \textcolor{white}{0}680 \\ 
        \midrule
        Gestures & meaningful body poses & *G & 876 \\ 
        & that non-signers also do & & \\
        \midrule 
        Fingerspelling & signing letters of the BSL & *FS & 1340 \\ 
        & alphabet to spell a word & & \\
        \midrule
        Depicting signs & signs which describe shape, & *D & 2569 \\ 
        & size, movement or handling & & \\  
        \midrule
        Pointing & sign with index finger & *P & (2477 *P1) 4179 \\ 
        & or flat hand &  & \\
        \midrule 
        Total & & & \textcolor{white}{00}11K \\ 
        \bottomrule
        \end{tabular}
            \vspace{-0.3cm}
	    \caption{\textbf{Sign types:} We present the different sign type annotations collected and their occurrence in \testCSLR. We also provide a brief description of each sign type and their annotation short form. See Fig.~\ref{fig:app:star_signs_qualitative_egs} for visual examples. Note we add in parenthesis, the subset of pointing annotations referred to as *P1 in Section~\ref{appendix:stats}, to approximate the linguistic pointing such as pronominal referencing.
	}
	\label{tab:app:sign-types}
\end{table*}

\noindent \textbf{Sign types.} As mentioned in \if\sepappendix1{Sec.~4}
\else{Sec.~\ref{sec:data}}
\fi
of the main paper, in addition to lexical signs that have an English word equivalent,
a wide range of sign types are annotated.
This is mainly motivated by signs that cannot easily be associated to lexical words:
pointing signs or fingerspelled signs are such examples.

\begin{figure*}
    \centering
    \includegraphics[width=\linewidth]{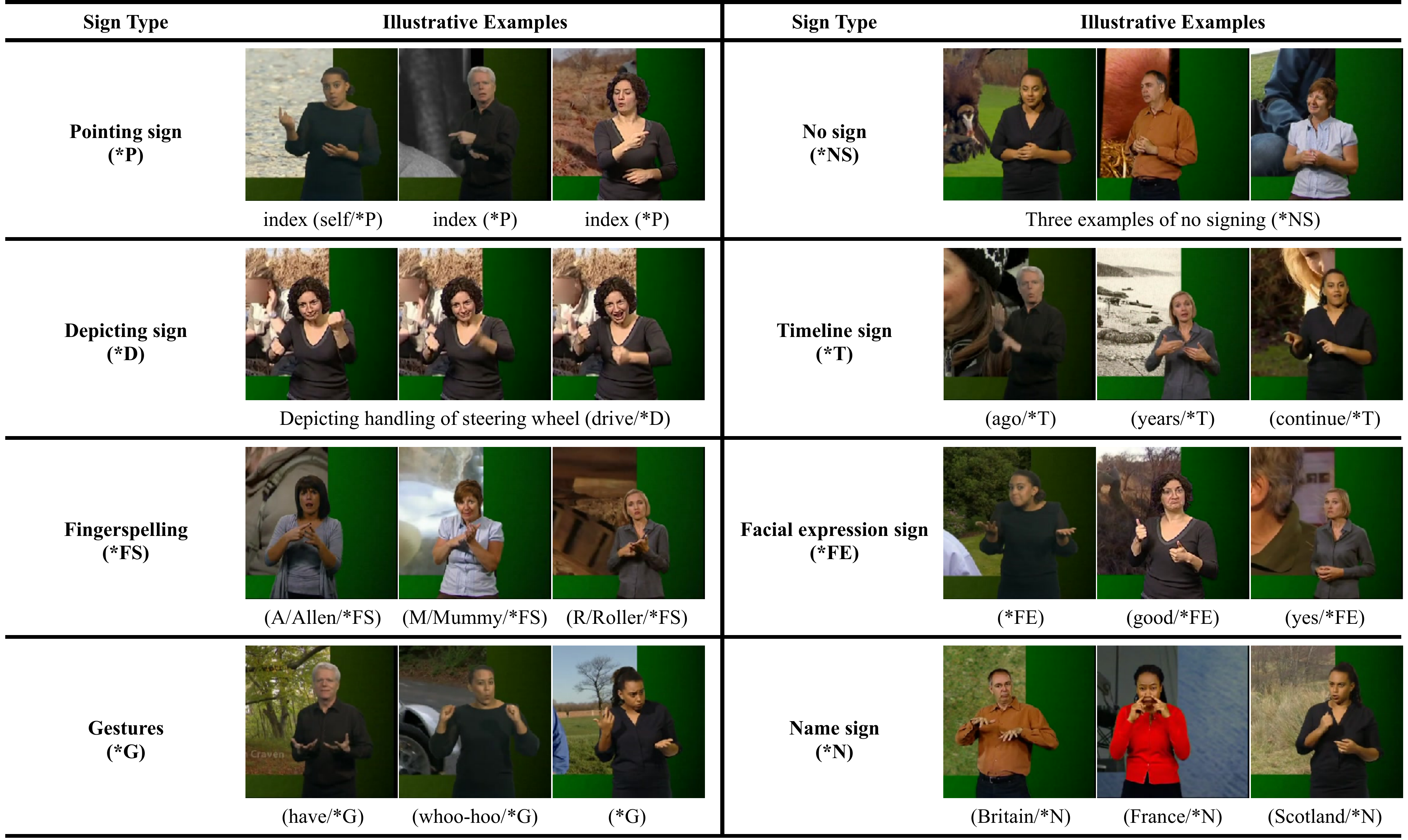}
    \vspace{-0.6cm}
    \caption{\cvprmodifs{\textbf{Sign type examples:} For each sign type annotation in Tab.~\ref{tab:app:sign-types}, we show 3 illustrative samples.} 
    }
    \label{fig:app:star_signs_qualitative_egs}
\end{figure*}

In Tab.~\ref{tab:app:sign-types}, we present the different sign type annotations collected and their occurrence in \testCSLR.
We note that for the CSLR evaluation, we only use signs that are associated to words (or phrases), filtering out the non-lexical special sign types since these are outside the scope of this work. We highlight that unknown, i.e.\ unrecognisable signs denoted by *U, are also filtered out for the CSLR evaluation since they 
are not associated to a lexical word. 
In Fig.~\ref{fig:app:star_signs_qualitative_egs}, we show several examples per sign type. 

We are also aware that our *P pointing annotations' definition differs from standard pointing as seen in the linguistic literature (e.g.\ BSL-Corpus~\cite{bslcorpus17}).
Annotators are encouraged to mark lexical signs that
contain the pointing \textit{handshape} (such as the sign
for `rocket' which is signed with index finger pointing upwards) even when
they are not pronominal reference~\cite{cormier2013}.
This can potentially be beneficial 
in future work to recognise the role of pointing in context.
Our pointing annotations are therefore a superset of the linguistic definition.
We approximate the linguistic pointing annotations containing locative pointing or pronominal reference~\cite{cormier2013} (shown in parenthesis in Table~\ref{tab:app:cslr-stats}) by counting pointing annotations 
(i) without any lexical word associated to it or
(ii) annotations associated to words in the following list: `me', `this', `I', `here', `in', `that', `you'.
These words were determined by inspecting the most frequent words assigned to *P signs. We refer to this subset of *P signs as *P1, and the rest as *P2.

\begin{figure*}
    \centering
    \begin{minipage}[c]{.33\linewidth}
        \centering
        \includegraphics[width=\linewidth]{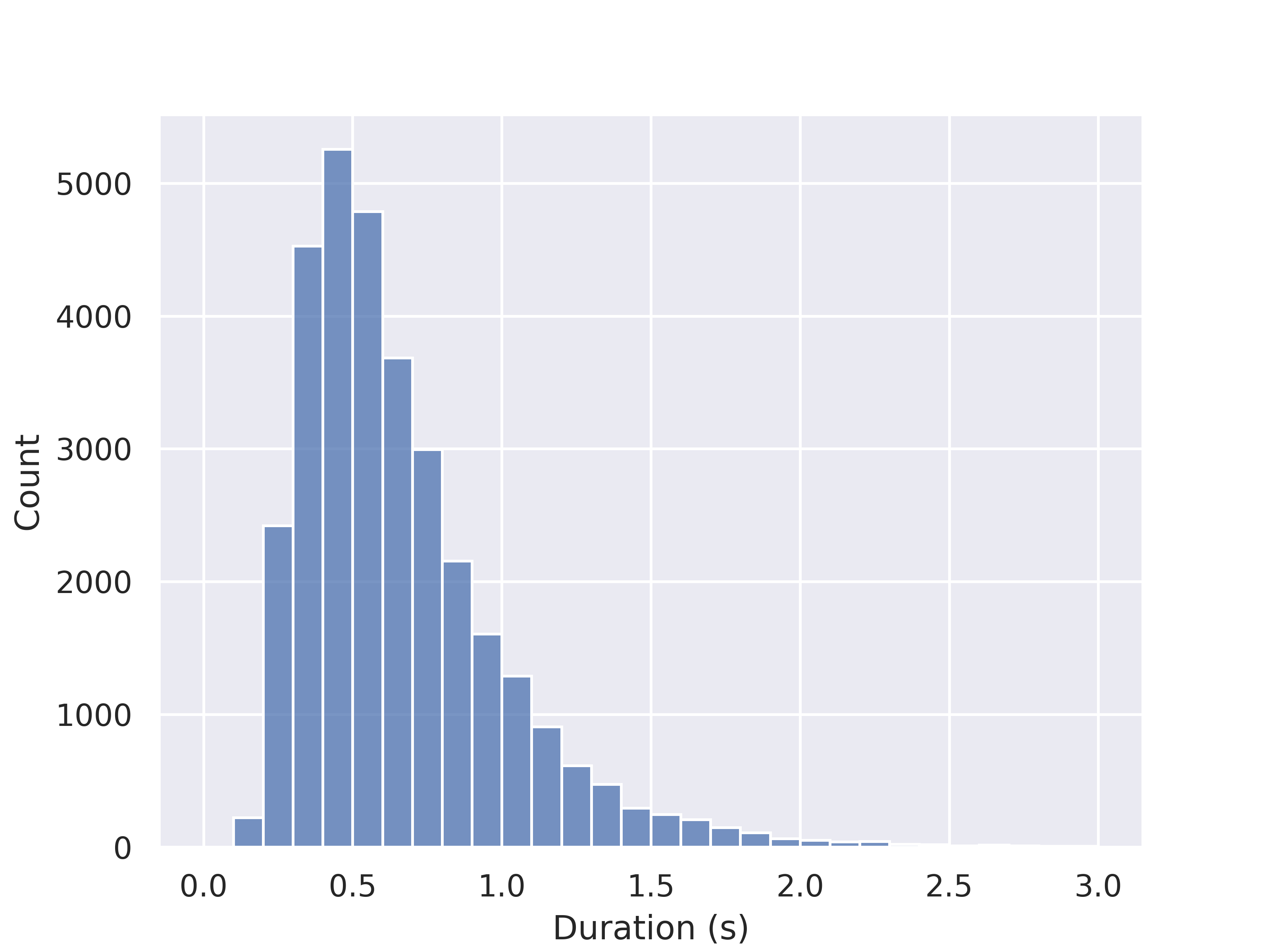}
    \end{minipage}
    \begin{minipage}[c]{.33\linewidth}
        \centering
        \includegraphics[width=\linewidth]{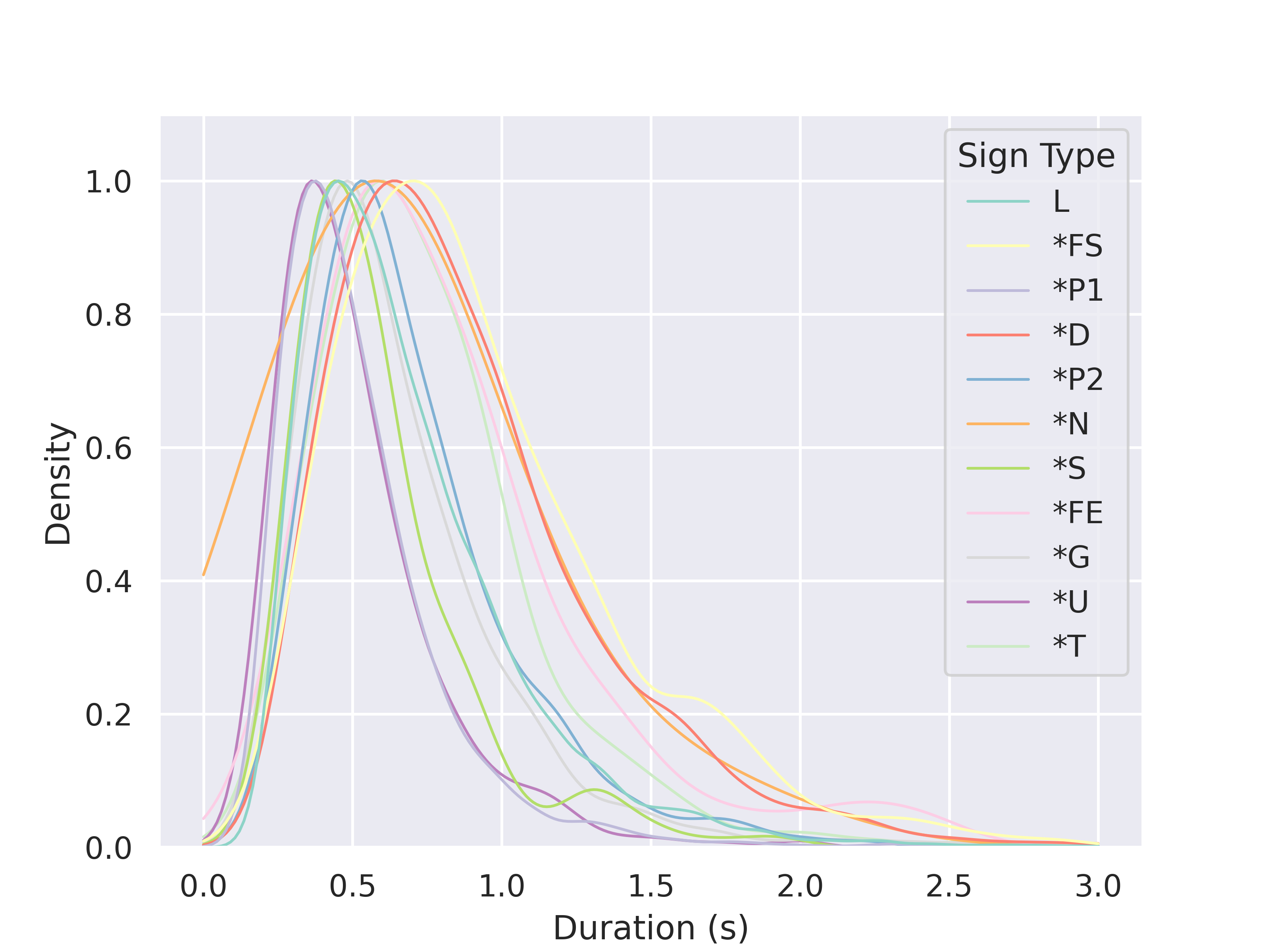}
    \end{minipage}
    \begin{minipage}[c]{.33\linewidth}
        \centering
        \includegraphics[width=\linewidth]{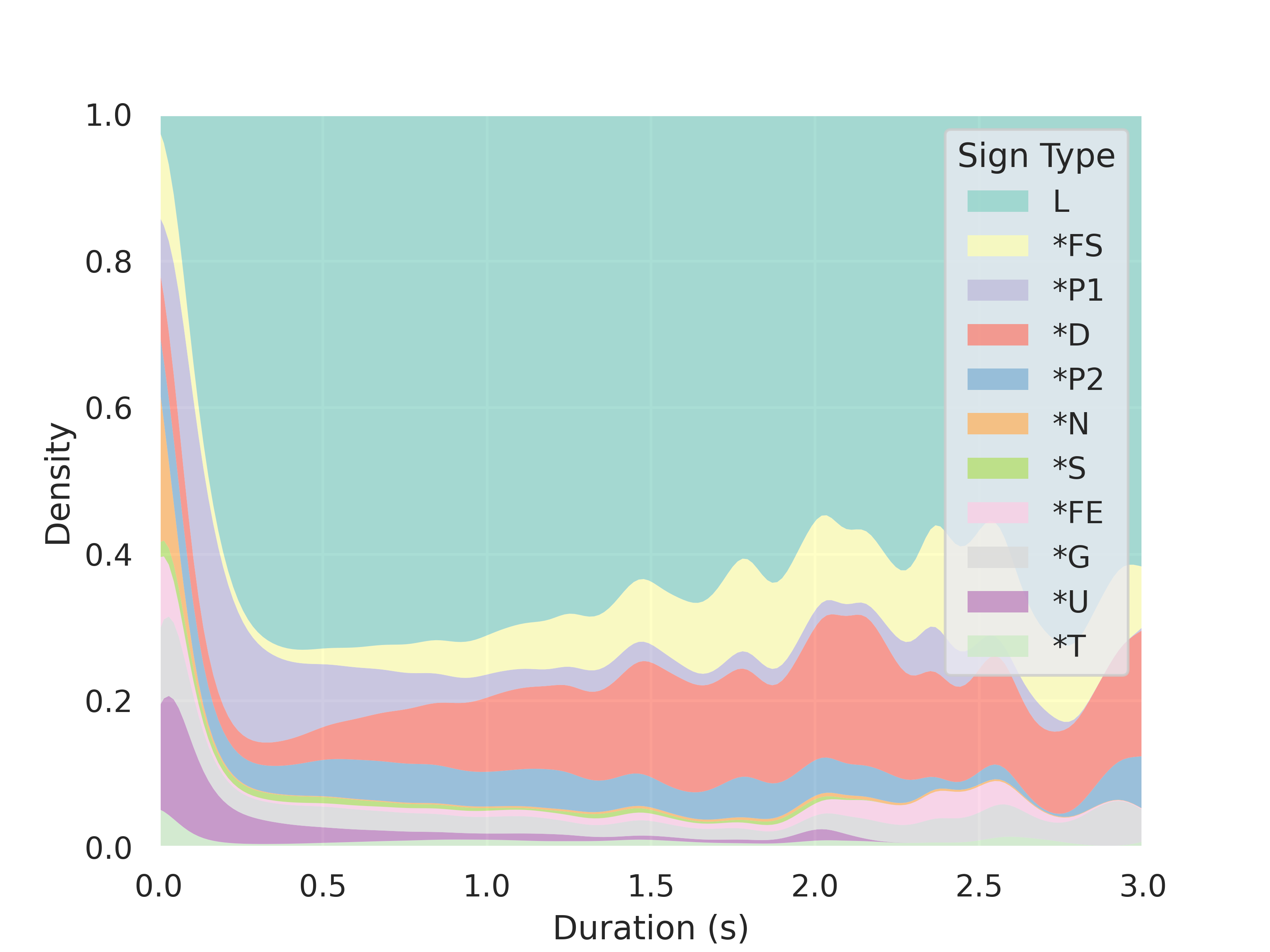}
    \end{minipage}
    \vspace{-0.3cm}
    \caption{\textbf{Statistics on \testCSLR annotations' duration:} We show \textbf{(left)} the histogram of durations for all annotations; \textbf{(middle)} the kernel density estimation of duration for each annotation type; and \textbf{(right)} the distribution over annotation types for different durations. L denotes purely lexical signs. See text for interpretation.
    }
    \label{fig:app:duration_hist}
\end{figure*}

\noindent \textbf{Annotation duration.} In Fig.~\ref{fig:app:duration_hist}, we show three plots on the duration of \testCSLR annotations. 
First, the left plot shows the histogram of durations: we see that a large portion of the annotations last between 0.4 and 0.6 seconds, which is in accordance to the estimates of previous works~\cite{Buehler08}. 
Second, the middle plot shows the duration distribution per sign type; we see that fingerspelled signs (*FS) last longer than other sign types, potentially because fingerspelling involves multiple handshapes (one per letter) to express a word. On the other hand, pointing signs (*P1) seem to have shorter duration than others.
Third, the right plot shows the proportion of sign types per duration, and we see that purely lexical signs dominate other annotation types. 
We also observe that the proportion of fingerspelled and depicting signs increases with longer duration.

\section{Experiments}\label{section:app-exp}

\subsection{CSLR baselines}
\label{appendix:cslr_baselines}

In Tab.~\ref{tab:app:cslr_baselines}, we provide a more detailed version of
\if\sepappendix1{Tab.~4}
\else{Tab.~\ref{tab:cslr_retrieval}}
\fi
of the main paper.
In particular, we give results with/without our post-processing filtering $\dagger$ on ISLR-based methods from \cite{Albanie21b,Momeni22,Prajwal22a}, and show the impact of thresholds on spotting-based methods from \cite{Albanie21b,Momeni22} that use subtitles. We see that without the post-processing heuristics, the ISLR-based methods and our CSLR-based method do not obtain reasonable results, mainly due to the noise from many false positives. Our model performs the best both in the settings (i) with and (ii) without post-processing, amongst all the approaches that do not use subtitle-cues. 

\begin{table*}
    \centering
    \setlength{\tabcolsep}{7pt}
    \resizebox{.65\linewidth}{!}{
        \begin{tabular}{lccc}
            \toprule 
            & WER $\downarrow$ & mIOU $\uparrow$ \\
            \midrule
            Subtitle-based auto.\ annots.~\cite{Albanie21b} [M$_{.5}$D$_{.7}$A$_{.0}$] & 115.8 & 13.0 \\
            Subtitle-based auto.\ annots.~\cite{Albanie21b} [M$_{.8}$D$_{.8}$] & \textcolor{white}{0}93.8 & \textcolor{white}{0}7.1 \\
            \midrule
            Subtitle-based auto.\ annots.~\cite{Momeni22} [M$^{*}_{.8}$D$^{*}_{.8}$A$_{.0}$P$_{.5}$E N] & \textcolor{white}{0}90.7 & 20.0 \\
            Subtitle-based auto.\ annots.~\cite{Momeni22} [M$^{*}_{.8}$D$^{*}_{.8}$A$_{.0}$P$_{.5}$ ]& \textcolor{white}{0}85.5 & 18.1 \\
            Subtitle-based auto.\ annots.~\cite{Momeni22} [M$^{*}_{.7}$D$^{*}_{.9}$A$_{.4}$P$_{.2}$]  & \textcolor{white}{0}85.9 & 15.3 \\
            \midrule
            ISLR I3D-2K~\cite{Albanie21b} & 453.5 & \textcolor{white}{0}8.6 \\
            ISLR I3D-2K~\cite{Albanie21b} $\dagger$ & \textcolor{white}{0}82.5 & 18.1 \\
            \midrule
            ISLR I3D-8K~\cite{Momeni22} & 343.6 & 14.9 \\
            ISLR I3D-8K~\cite{Momeni22} $\dagger$ & \textcolor{white}{0}74.7 & 27.0 \\
            \midrule
            ISLR Swin-8K~\cite{Prajwal22a}  & 293.9 & 17.6 \\
            ISLR Swin-8K~\cite{Prajwal22a} $\dagger$ & \textcolor{white}{0}71.7 & 30.1 \\
            \midrule 
            \rowcolor{aliceblue} \modelname (Ours) & 201.4 & 22.4 \\
            \rowcolor{aliceblue} \modelname (Ours)$\dagger$ & \textcolor{white}{0}\textbf{65.5} & \textbf{35.5} \\
        \bottomrule
        \end{tabular}
    }
        \vspace{-0.3cm}
	\caption{\textbf{Analysis on CSLR baselines:}
	$\dagger$ denotes the optimal subset of the annotations that gives the highest performance. 
	The values without $\dagger$ are obtained using the raw model predictions and collapsing continuous repetitions. We observe that performing post-processing is essential to obtain reasonable results. Our model gives a clear improvement over the current best CSLR method.
	}
	\label{tab:app:cslr_baselines}
        \vspace{-0.3cm}
\end{table*}

\subsection{Oracle CSLR performance}
\label{appendix:oracle}
We conduct an oracle experiment to determine an upper-bound for: ``What is the best performance we can obtain if we remove all the false positives predictions from our model after post-processing?". To do this, we compare the post-processed predictions of our best model with the words in the ground-truth sequence and remove the ones that are false positives. In Tab.~\ref{tab:app:oracle}, we see that we can obtain a large improvement in scores, but the WER is still relatively high. This indicates that our model should recognise more signs in the sequences for it to lower the WER even more.

\begin{table}
    \centering
    \setlength{\tabcolsep}{5pt}
    \begin{tabular}{lcc}
            \toprule 
            & WER $\downarrow$ & mIOU $\uparrow$ \\
            \midrule
            \modelname (Ours) & 65.5 & 35.5 \\
            \midrule
            Oracle & 61.0 & 42.3 \\
        \bottomrule
        \end{tabular}
            \vspace{-0.3cm}
	    \caption{\textbf{Oracle analysis:} We compute oracle results by removing all false positives (i.e.\ words in the post-processed predicted sequence that are not in the ground-truth sequence). We see that despite removing false positives, the WER is still relatively high, implying that we either miss words in the ground-truth sequence or we predict them at the incorrect location.
		}
		\label{tab:app:oracle}
\end{table}

\subsection{CSLR performance breakdown and analysis}
\cvprmodifs{
In Tab~\ref{tab:app:clsr_breakdown}, we report the CSLR performance breakdown 
by excluding specific sign types from both ground truth and prediction \textit{timelines},
i.e.\ removing corresponding frames.
We further examine the oracle performance
by assuming we perfectly recognise the words
that correspond to certain sign types,
such as pointing and fingerspelling.
Specifically, we investigate the impact of 
(i) all * annotations, (ii) pointing *P annotations, and (iii) fingerspelling *FS annotations.
We see that in an ideal scenario where pointing or fingerspelled annotations are perfectly predicted,
the overall CSLR performance is substantially improved,
leading to a significant WER reduction of $-14.8$ (65.5 vs 50.7), and even more if all * sign types are correctly recognised (49.3).
}

\begin{table}
    \centering
    \setlength{\tabcolsep}{5pt}
    \resizebox{1\linewidth}{!}{
    \begin{tabular}{l|cc|cc}
        \toprule 
        & \multicolumn{2}{c|}{Filter} & \multicolumn{2}{c}{Oracle} \\
        Sign Type & WER $\downarrow$ & mIOU $\uparrow$ & WER $\downarrow$ & mIOU $\uparrow$ \\
        \midrule
        \modelname (Ours) & 65.5 & 35.5 & - & - \\
        \midrule
        *FS & 64.9 & 36.5 & 54.4 & 48.0 \\
        *P & 65.0 & 35.7 & 56.0 & 47.3 \\
        *P/*FS & 64.2 & 36.9 & 50.7 & 50.6 \\
        all * & 64.3 & 37.0 & 49.3 & 51.7 \\
        \bottomrule
    \end{tabular}
    }
        \vspace{-0.3cm}
	\caption{
            \cvprmodifs{
            \textbf{CSLR performance breakdown:}
            We experiment with 
            (i)~filtering out certain annotation types from both ground truth and prediction timelines (\textbf{Filter} column), and 
            (ii)~replacing predictions with ground truth labels for these annotations (\textbf{Oracle} column).
            We observe that pointing and fingerspelling, in particular, contribute to a significant WER reduction (-14.8).
    	}
        }
	\label{tab:app:clsr_breakdown}
\end{table}

\section{Ablations}\label{section:app-ablations}

In this section, we provide complementary ablations to what is done in 
\if\sepappendix1{Sec.~5.3}
\else{Sec.~\ref{sec:experiments}.}
\fi
of the main paper.
\cvprmodifs{
We first look at the impact of changing the frozen text backbone, from pre-trained T5 model to pre-trained MPNet~\cite{song2020mpnet}, on both CSLR and retrieval (Sec.~\ref{appendix:text_encoder}).}
We then show additional retrieval results obtained on 20k unseen \testManual subtitles (Sec.~\ref{appendix:retrieval_baselines}).
\cvprmodifs{Next, we show the impact of evaluating with synonyms at test time on CSLR performance (Sec.~\ref{appendix:syns}).} 
We then provide ablations on: 
(i) loss weights (Sec.~\ref{appendix:loss_weight}),
(ii) and subtitle alignment (Sec.~\ref{appendix:alignment}).

\subsection{Text encoder}
\label{appendix:text_encoder}

\cvprmodifs{
In Tab.~\ref{tab:app:text_encoder}, we experiment with replacing the frozen T5~\cite{raffel2020t5} text encoder with MPNet~\cite{song2020mpnet}.
We first see that opting for T5 instead of MPNet yields performance gains on all reported metrics.
Moreover, we observe that our joint training helps, for both text encoders, improving over (i) the strong ISLR Baseline for CSLR, as well as (ii) the sole retrieval training overall.
}

\begin{table*}
    \centering
    \setlength{\tabcolsep}{5pt}
    \resizebox{1\linewidth}{!}{
        \begin{tabular}{c|c|c|ccccc|ccc|ccc} 
            \toprule 
            & & & \multicolumn{5}{c|}{CSLR} & \multicolumn{3}{c|}{\textbf{T2V}} & \multicolumn{3}{c}{\textbf{V2T}} \\ 
            \textbf{Text Backbone} & \textbf{SentRet} & \textbf{SignRet} & WER $\downarrow$ & mIoU $\uparrow$ & \multicolumn{3}{c|}{F1@$\{0.1, 0.25, 0.5\} \uparrow$} & R@1 $\uparrow$ & R@5 $\uparrow$ & R@10 $\uparrow$ & R@1 $\uparrow$ & R@5 $\uparrow$ & R@10 $\uparrow$ \\
            \midrule
            \rowcolor{goodcolor} \multicolumn{3}{l|}{ISLR Baseline} & 71.7 & 30.1 & 40.0 & 37.9 & 27.2 & - & - & - & - & - & - \\
            \midrule
            MPNet & \cmark & \xmark & - & - & - & - & -& 44.7 & 64.9 & 70.6 & 43.0 & 64.4 & 69.9 \\
            T5 & \cmark & \xmark & - & - & - & - & - & 50.5 & 69.5 & 75.1 & 49.7 & \textbf{69.7} & 74.7 \\
            \midrule
            MPNet & \cmark & \cmark & 66.3 & 34.9 & 46.4 & 45.3 & 36.3 & 46.1 & 66.2 & 72.2 & 44.3 & 65.1 & 71.8 \\
            T5 & \cmark & \cmark &  \textbf{65.5} & \textbf{35.5} & \textbf{47.1} & \textbf{46.0} & \textbf{37.1} & \textbf{51.7} & \textbf{69.9} & \textbf{75.4} & \textbf{50.2} & 69.1 & \textbf{74.7} \\
        \bottomrule
        \end{tabular}
    }
    \vspace{-0.3cm}
	\caption{
    \cvprmodifs{
    \textbf{Impact of the text encoder:}
	We experiment with the choice of the text encoder in our model. More specifically, we compare T5 (\texttt{t5-large})~\cite{raffel2020t5} and MPNet (\texttt{all-mpnet-v2})~\cite{song2020mpnet}.
        We observe gains on all reported metrics by opting for T5 over MPnet. 
	}
    }
	\label{tab:app:text_encoder}
\end{table*}

\subsection{Retrieval on \testManual}
\label{appendix:retrieval_baselines}

In \if\sepappendix1{Tab.~3}
\else{Tab.~\ref{tab:retrieval_ablations},}
\fi 
of the main paper, we provide ablations by measuring performance on 2K \valManual sentences.
In Tab.~\ref{tab:app:retrieval_ablations},
we report the same ablations 
on 20K \testManual sentences.
The same conclusions hold on this more challenging
(larger) retrieval benchmark.

\begin{table*}
    \centering
    \setlength{\tabcolsep}{5pt}
    \resizebox{1\linewidth}{!}{
        \begin{tabular}{cccc|ccccc|ccccc} 
            \toprule 
            & & & & \multicolumn{5}{c|}{\textbf{T2V}} & \multicolumn{5}{c}{\textbf{V2T}} \\ 
            \textbf{Pool} & \textbf{Text Encoder} & \textbf{SentRet} & \textbf{Sign-level loss} & R@1 $\uparrow$ & R@5 $\uparrow$ & R@10 $\uparrow$ & R@50 $\uparrow$& MedR $\downarrow$ & R@1 $\uparrow$ & R@5 $\uparrow$ & R@10 $\uparrow$ & R@50 $\uparrow$& MedR $\downarrow$ \\ 
            \midrule
            \rowcolor{goodcolor}\texttt{cls} & MPNet & InfoNCE & \xmark & 16.1 & 29.5 & 35.8 & 51.1 & 45 & 14.5 & 27.8 & 34.1 & 50.1 & 50 \\ 
            \rowcolor{goodcolor}\texttt{cls} & T5-large & InfoNCE & \xmark & 19.5 & 35.1 & 41.8 & 58.1 & 24 & 18.9 & 33.8 & 40.7 & 57.4 & 25 \\
            \rowcolor{goodcolor}\texttt{max} & T5-large & InfoNCE & \xmark & 21.9 & 36.9 & 43.6 & 59.0 & 20 & 20.7 & 36.0 & 42.8 & 58.8 & 21 \\
            \rowcolor{goodcolor}\texttt{cls} & T5-large & HN-NCE & \xmark & 25.8 & 42.1 & 48.9 & 63.0 & 12 & 27.2 & 43.0 & 49.2 & 63.7 & 11 \\
            \rowcolor{goodcolor}\texttt{max} & T5-large & HN-NCE & \xmark & 28.6 & 44.2 & 50.5 & 64.8 & 10 & 27.5 & 43.6 & 50.3 & 64.5 & 10 \\
            \rowcolor{aliceblue} \texttt{max} & T5-large & HN-NCE & CE & 28.5 & 44.4 & 50.8 & 64.8 & 10 & 27.9 & 43.9 & 50.4 & 64.5 & 10 \\
            \rowcolor{aliceblue} \texttt{max} & T5-large & HN-NCE & SignRet & \textbf{29.4} & \textbf{45.2} & \textbf{51.5} & \textbf{64.9} & \textcolor{aliceblue}{0}\textbf{9} & \textbf{28.1} & \textbf{44.9} & \textbf{51.0} & \textbf{64.9} & \textcolor{aliceblue}{0}\textbf{9} \\
        \bottomrule
        \end{tabular}
    }   
        \vspace{-0.3cm}
	\caption{\textbf{Retrieval ablations on \testManual:}
	We investigate several design choices
        similar to
	\if\sepappendix1{Tab.~3}
	\else{Tab.~\ref{tab:retrieval_ablations}}
	\fi 
 of the main paper,
 but this time on the larger gallery of \testManual (instead of \valManual). Please refer to
	\if\sepappendix1{Sec.~3.1}
	\else{Sec.~\ref{subsec:objectives}}
	\fi 
	of the main paper for a description of this experiment along with the interpretation of the results.
	}
	\label{tab:app:retrieval_ablations}
\end{table*}

\subsection{Synonyms for CSLR}
\label{appendix:syns}
We have shown through a series of qualitative examples
the importance of having a more comprehensive synonyms list in the evaluation process. 
Here, we look at the impact of considering synonyms as correct when evaluating for CSLR in terms of performance.
In Tab.~\ref{tab:app:syns}, we observe major gains on all reported metrics, when considering synonyms as correct,  despite our synonyms list being far from perfect as explained in our qualitative analysis in the main paper.

\begin{table}
    \centering
    \setlength{\tabcolsep}{5pt}
    \resizebox{1.0\linewidth}{!}{
    \begin{tabular}{c|c|ccccc}
        \toprule 
        & & \multicolumn{5}{c}{\textbf{CSLR}} \\
        Model & Synonyms & WER $\downarrow$ & mIOU $\uparrow$ & \multicolumn{3}{c}{F1@$\{0.1, 0.25, 0.5\} \uparrow$} \\
        \midrule
        \rowcolor{goodcolor}\textsc{ISLR Baseline} & \xmark & 78.0 & 22.6 & 32.6 & 30.9 & 21.6 \\
        \rowcolor{goodcolor}\textsc{ISLR Baseline} & \cmark & 71.7 & 30.1 & 40.0 & 37.9 & 27.2 \\
        \midrule
        \modelname (Ours) & \xmark & 73.3 & 26.5 & 37.8 & 36.8 & 29.4 \\
        \modelname (Ours) & \cmark & \textbf{65.5} & \textbf{35.5} & \textbf{47.1} & \textbf{46.0} & \textbf{37.1} \\
        \bottomrule
    \end{tabular}
    }
        \vspace{-0.3cm}
	\caption{\textbf{Impact of synonyms at test time:}
        We observe major gains on all reported metrics when considering synonyms as correct for CSLR evaluation.
        This shows that having a precise and exhaustive list of synonyms is 
        important for CSLR.	
	}
	\label{tab:app:syns}
\end{table}

\subsection{Weighting of losses on CSLR performance}
\label{appendix:loss_weight}
\begin{figure}
    \centering
    \includegraphics[width=\linewidth]{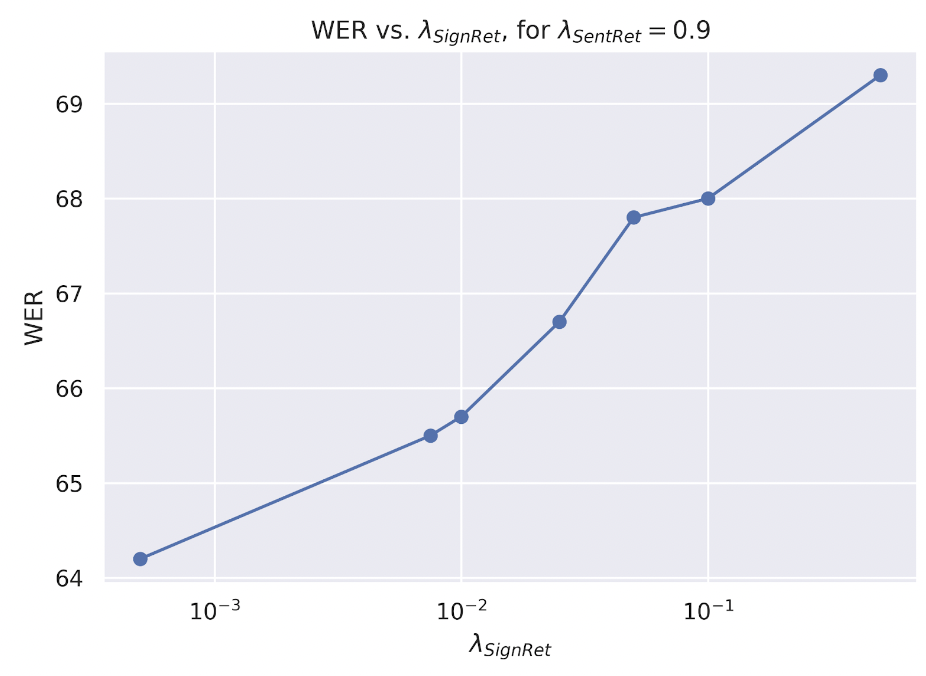}
    \vspace{-1.0cm}
    \caption{\textbf{Weighting of losses on CSLR performance:} We look at the impact of changing $\lambda_{\text{SignRet}}$ for fixed $\lambda_{\text{SentRet}} = 0.9$. We observe that WER decreases with lower $\lambda_{\text{SignRet}}$ values.}  
    \label{fig:app:loss_weights}
\end{figure}

We study the impact of loss weighting on CSLR performance. 
More specifically, in Fig.~\ref{fig:app:loss_weights}, 
we plot WER performance obtained by jointly training with our two losses (SentRet and SignRet) 
against varying loss weights for the sign-level loss, 
fixing the weight given to the sentence-level loss.
Surprisingly, we observe that giving less weight to the CSLR-relevant SignRet loss actually brings improvement to CSLR.

We see that WER decreases with lower $\lambda_{\text{SignRet}}$ values.
A potential explanation could be
that the SentRet loss regularizes
the training, as we observed SignRet
alone is prone to overfitting.

\begin{table*}
    \centering
    \setlength{\tabcolsep}{5pt}
    \resizebox{0.7\linewidth}{!}{
    \begin{tabular}{l|ccccc|cccc}
        \toprule 
         & \multicolumn{5}{c|}{\textbf{\testCSLR}} & \multicolumn{2}{c}{\textbf{T2V \valManual}} & \multicolumn{2}{c}{\textbf{V2T \valManual}} \\
        Subtitles & WER $\downarrow$ & mIOU $\uparrow$ & \multicolumn{3}{c|}{F1@$\{0.1, 0.25, 0.5\}$ $\uparrow$} & R@1 $\uparrow$ & R@5 $\uparrow$ & R@1 $\uparrow$ & R@5 $\uparrow$  \\
        \midrule
        Audio-aligned & 68.0 & 33.5 & 44.5 & 43.3 & 33.5 & 25.2 & 43.8 & 27.7 & 44.2 \\
        Signing-aligned~\cite{Bull21} & \textbf{65.5} & \textbf{35.5} & \textbf{47.1} & \textbf{46.0} & \textbf{37.1} & \textbf{51.7} & \textbf{69.9} & \textbf{50.2} & \textbf{69.1}\\
        \bottomrule
    \end{tabular}
    }
        \vspace{-0.3cm}
	\caption{\textbf{Subtitle alignment:} Using automatic signing-aligned subtitles~\cite{Bull21} instead of audio-aligned subtitles drastically improves retrieval performance, and moderately improves CSLR performance.
	}
	\label{tab:app:sub_align}
\end{table*}

\subsection{Subtitle alignment}
\label{appendix:alignment}
In Tab.~\ref{tab:app:sub_align}, we measure the impact of using audio-aligned subtitles instead of automatic signing-aligned subtitles \cite{Bull21}. We observe a notable gain in retrieval performance using automatic signing-aligned subtitles (R@1 for both T2V and V2T almost doubles). This is expected as audio-aligned subtitles are not necessarily aligned with the signed content in videos, but instead with the audio soundtrack accompanying the original TV broadcasts. We also observe a drop in CSLR performance when using audio-aligned subtitles.

\begin{figure*}
    \centering
    \includegraphics[width=\linewidth]{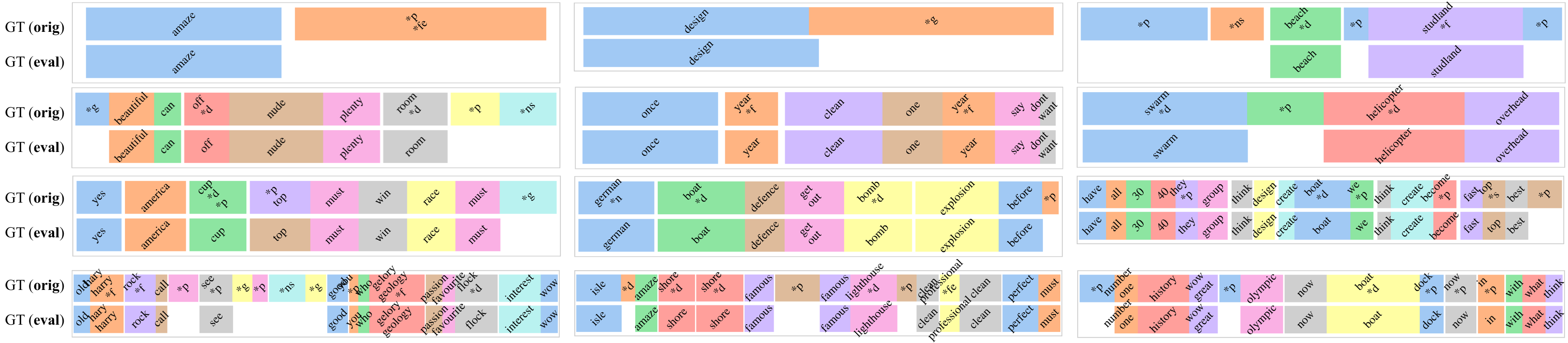}
    \vspace{-0.3cm}
    \caption{\textbf{Ground truth processing:} 
    We provide examples of the original ground truth annotations as collected by annotators (GT orig), and the version we use at evaluation time (GT eveal).
    }
    \label{fig:app:raw_vs_eval_gt}
\end{figure*}

\begin{figure*}
    \centering
    \includegraphics[width=\linewidth]{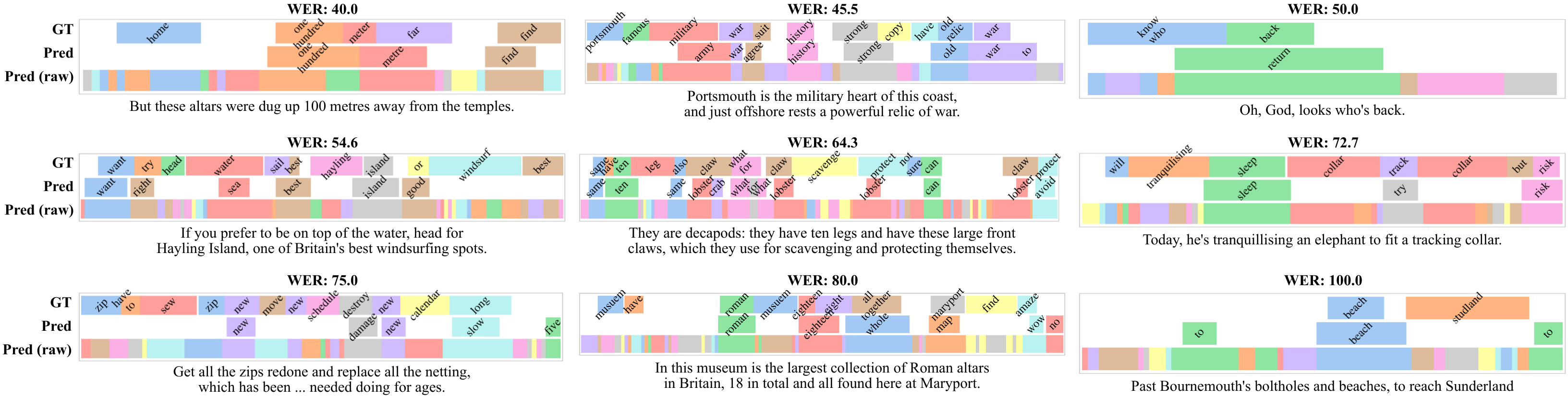}
    \vspace{-0.3cm}
    \caption{\textbf{Effect of post-processing the predictions:} 
    We show that processing the frame-level sign recognitions into segment-level ones (as explained in
    \if\sepappendix1{Sec.~3.3}
    \else{Sec.~\ref{subsec:supervision}}
    \fi
    of the main paper)
    effectively cleans up a large portion of the noise. For each sample, top row corresponds to the ground truth (GT), middle row to post-processed CSLR predictions from \modelname model, and bottom to unfiltered raw predictions.
    }
    \label{fig:app:post_processing}
\end{figure*}

\section{Qualitative Examples}\label{section:app-qual-results}

\subsection{CSLR}\label{subsec:app:qual_cslr}

\noindent \textbf{Ground truth processing.}
We show in Fig~\ref{fig:app:raw_vs_eval_gt} several sequences 
illustrating how we obtain the ground truth for evaluation, in particular with the removal of * annotations and keeping only the words associated to each annotation.

\noindent \textbf{Effect of post-processing the predictions.}
We show in Fig~\ref{fig:app:post_processing} qualitative results
with and without the post-processing procedure described in
\if\sepappendix1{Sec.~3.3}
\else{Sec.~\ref{subsec:supervision}}
\fi
of the main paper. These examples particularly highlight the importance of the post-processing we propose in our method.

\begin{figure*}
    \centering
    \includegraphics[width=\linewidth]{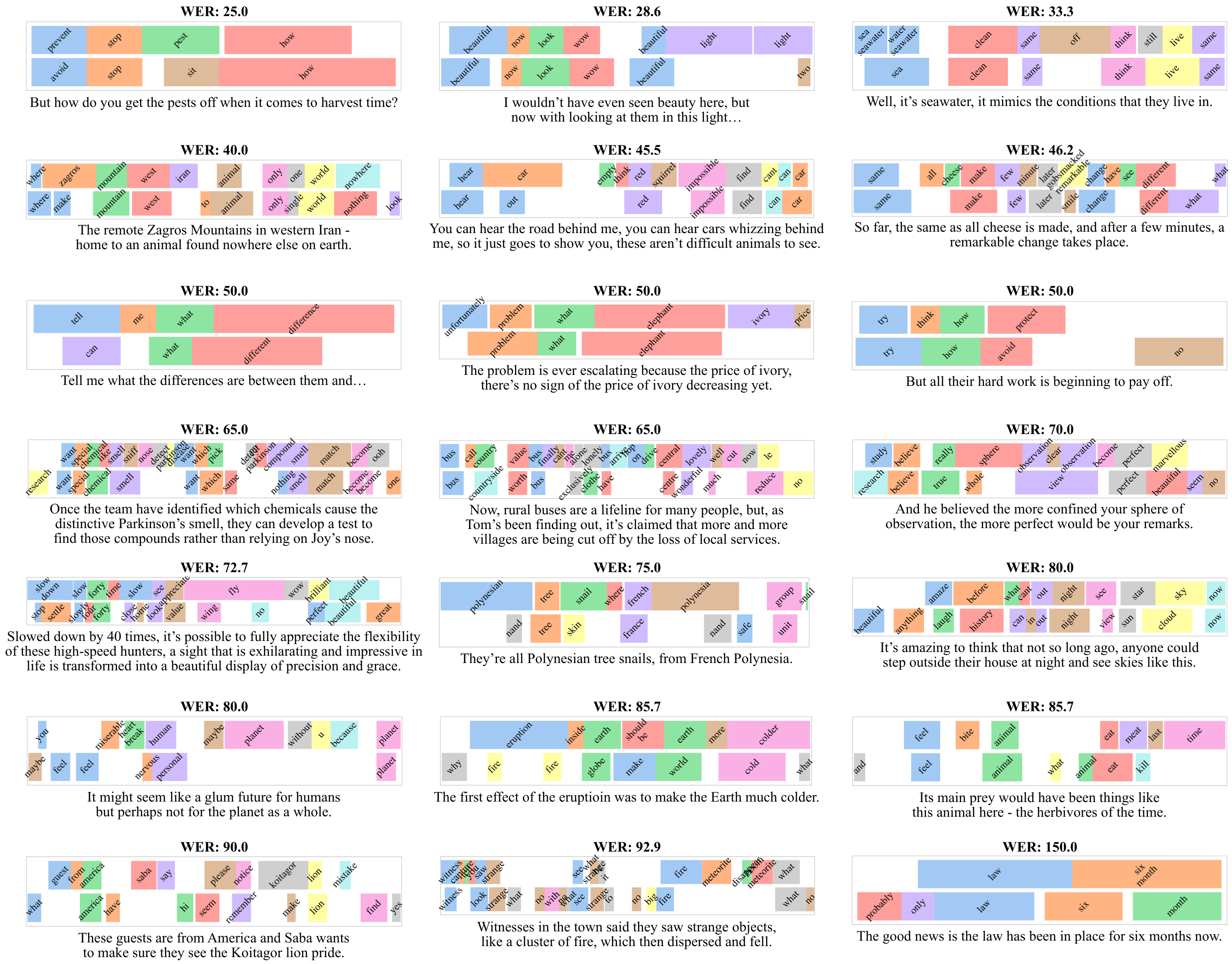}
    \vspace{-0.3cm}
    \caption{\textbf{Additional qualitative CSLR results.} We provide additional examples for our model's CSLR predictions (bottom row per sample) on \testCSLR, as well as the the ground truth (top row per sample). As in 
    \if\sepappendix1{Fig.~4}
    \else{Fig.~\ref{fig:qualitative}}
    \fi
    of the main paper, we display WER for each example, along with the corresponding subtitle to provide context for the reader.
    }
    \label{fig:app:cslr_timelines}
\end{figure*}

\noindent \textbf{Additional results.}
We show qualitative examples of CSLR performance in Fig.~\ref{fig:app:cslr_timelines}.
For each signing video sample, we show the ground truth and predicted glosses, along with the English subtitle.
We also show the WER and IOU per sample.
We observe that our model is able to correctly predict an important proportion of the ground truth annotations.

\subsection{Retrieval}\label{subsec:app:qual_ret}
We show qualitative results of retrieval performance both for V2T and T2V settings in Tab.~\ref{tab:app:v2t} and Tab.~\ref{tab:app:t2v}, respectively. For ease of visualisation, we report the subtitles corresponding to the sign video V. We note that the retrieved subtitle (in the V2T case) or retrieved video subtitles (in the T2V case) are very often in the Top-5 and that other top retrieved results are often semantically similar to the correct result (as shown by similar words in \textcolor{blue}{blue}), highlighting the strong capabilities of our joint sign-video-to-text embedding space.

\begin{table*}
    \centering
    \setlength{\tabcolsep}{15pt}
    \resizebox{1.0\linewidth}{!}{
    \begin{tabular}{p{.5\linewidth}|p{.5\linewidth}}
        \toprule 
        Query Video (illustrated by its corresponding subtitle) & Retrieved Sentences \\
        \midrule
        She was deeply involved in farming, and her impassioned campaigning certainly ruffled a few feathers & \textbf{She was deeply involved in farming, and her impassioned campaigning certainly ruffled a few feathers. (0.74)} \\
        & To encourage them to do just that, \textcolor{blue}{farmers} have been  \textcolor{blue}{campaigning} hard (0.72) \\
        & She celebrated and  \textcolor{blue}{championed rural} life. (0.68) \\
        & And like Beatrix Potter, you have a \textcolor{blue}{real interest} in preserving the \textcolor{blue}{countryside}, especially \textcolor{blue}{farms}. (0.67) \\
        & Visit our website and tell us why they deserve to be our next \textcolor{blue}{farming} hero. (0.66) \\
        \midrule
        Like getting dressed up on a Friday night? & \textbf{Like getting dressed up on a Friday night? (0.73)} \\
        & But the homecoming celebration won't begin until much later this \textcolor{blue}{evening}. (0.69) \\
        & This is the festivities of the \textcolor{blue}{night}. (0.67) \\
        & \textcolor{blue}{Friday night} in... ..with just the BBC Two of us. (0.67) \\
        & It really does look furious \textcolor{blue}{tonight}. (0.66) \\
        \midrule
        But, actually, surprisingly they emit fewer & It's all just pure \textcolor{blue}{calcium carbonate}. (0.68) \\
        greenhouse gases, up to about 3.5.  & In the end, it came down to just \textcolor{blue}{three}. (0.66) \\
        & So, we look at all of these and bring them down into one unit, which we call kilograms of \textcolor{blue}{carbon dioxide} equivalent, which lets us make comparisons between different food chains. (0.66) \\
        & They left from North Wales on the \textcolor{blue}{3rd} of July, and they headed down the Irish Sea and then across the Bristol Channel, around Land's End, and they rendezvoused with the German tug the Gladiator off the Belgium coast. (0.65) \\
        & \textbf{But, actually, surprisingly they emit fewer greenhouse gases, up to about 3.5. (0.65)} \\
        \midrule
        And in the course of the afternoon, we'd start singing our choruses from Sunday school. & \textbf{And in the course of the afternoon, we'd start singing our choruses from Sunday school. (0.83)} \\ 
        & One would \textcolor{blue}{start} and then another group would join in and before long you'd have the whole beach \textcolor{blue}{singing}. (0.71) \\
        & Half an hour on \textcolor{blue}{Sunday} mornings. (0.70) \\
        & All the things you imagine you'd do on a British holiday on the beach. (0.67) \\
        & I remember, in Ireland, I stopped on the banks of Galway Bay and had a little \textcolor{blue}{sing}. (0.67) \\
        \midrule
        Now, I've got nice... marinana... marinara meatballs. & Yeah, I mean, you've probably got 20mm of fat there, and with the Mangalitza the fat is so creamy. (0.68) \\
        & Well, yes, if you're a fan of mussels, then, \textcolor{blue}{I think you're going to enjoy this}. (0.68) \\
        & \textbf{Now, I've got nice... marinana... marinara meatballs. (0.67)} \\
        & If you look through there, you can see Marum, not very well, but... (0.66) \\
        & I'm going with the chicken pesto pasta. (0.66) \\
\bottomrule
    \end{tabular}
    }
    \vspace{-0.3cm}
    \caption{\textbf{Qualitative V2T retrieval examples} obtained on the unseen \testManual sentences. To ease visibility, we report the subtitles corresponding to the query sign video, along with their the Top-5 retrieved sentences, as well as their similarities (in the range 0-1). Words in the retrieved sentence that are identical or 
    semantically similar to those in the query are shown in blue. Note the table continues on the next page.
    }
    \label{tab:app:v2t}
\end{table*}

\begin{table*}
    \centering
    \setlength{\tabcolsep}{15pt}
    \resizebox{1.0\linewidth}{!}{
    \begin{tabular}{p{.5\linewidth}|p{.5\linewidth}}
        \toprule 
         Query Video (illustrated by its corresponding subtitle) & Retrieved Sentences \\
        \midrule
        But keeping salmon contained in large numbers like this doesn't come without complications. & \textcolor{blue}{It does make it difficult}, but the current optimises the welfare of the \textcolor{blue}{salmon}. (0.74) \\
        & \textbf{But keeping salmon contained in large numbers like this doesn't come without complications. (0.72)}	\\
        & We're pioneers when it comes to \textcolor{blue}{farming in rough waters}, because it's probably the farming area where you \textcolor{blue}{farm salmon} with the strongest currents in the world. (0.68)	\\
        & \textcolor{blue}{The biggest threat to farmed salmon} around the world is sea lice, a marine parasite. (0.68) \\
        & We've got a little \textcolor{blue}{salmon} parr here, perfect. (0.67) \\
        \midrule 
        But it's a risk worth taking. & This is such a \textcolor{blue}{risky operation}! (0.70) \\
        & You know, \textcolor{blue}{it's exceedingly dangerous}. (0.69) \\
        & \textbf{But it's a risk worth taking. (0.67)} \\
        & It's going to be a \textcolor{blue}{huge gamble}. (0.67) \\
        & It is going to be a \textcolor{blue}{huge gamble}. (0.67) \\
        \midrule 
        We are looking forward to meet her and it's exciting because we've been working really hard to get the camp ready.	& I'm \textcolor{blue}{so excited to meet new people} every time because it gives me an opportunity to show them what we have here. (0.76)	\\
        & \textbf{We are looking forward to meet her and it's exciting because we've been working really hard to get the camp ready. (0.76)} \\
        & I'm \textcolor{blue}{particularly excited to be here} because it's not my first visit to Heartwood Forest. (0.74) \\
        & I've been coming to \textcolor{blue}{watch gatherings} more years than I care to remember, and \textcolor{blue}{I'm still just as excited}, and when you see a whole group of them break the skyline, galloping in towards you, all identical, it's fantastic. (0.73)	\\
        & I just feel like everything's kicking off, and \textcolor{blue}{it's quite exciting}. (0.71) \\
        \midrule 
        In the wild, they have to avoid being eaten by predators. & They spend most of their life \textcolor{blue}{trying to avoid being eaten by various things}. (0.71) \\
        & Some of \textcolor{blue}{these predators}, they rely on what they can find. (0.71) \\
        & \textcolor{blue}{They'll eat some of them, but they'll scare most away}, and that's why we end up like this. (0.70) \\
        & They get \textcolor{blue}{the animals}, then they're not sure where to graze them, then they think it's our duty to provide them with some grazing. (0.70) \\
        & \textbf{In the wild, they have to avoid being eaten by predators. (0.69)} \\
        \midrule
        Enjoy the grass and happy calving! & And when they are not rolling their balls down the hill \textcolor{blue}{our cattle are grazing it}. (0.72) \\
        & \textbf{Enjoy the grass and happy calving! (0.70)} \\
        & They're soon back on track and heading towards their \textcolor{blue}{summer grazing}. (0.69) \\
        & It's \textcolor{blue}{good to see them like this} - the way a white rhino must \textcolor{blue}{graze}. (0.69) \\
        & OK, for \textcolor{blue}{grazing}, you mean? (0.68) \\
        \bottomrule
    \end{tabular}
    }
    \label{tab:app:v2t_2}
\end{table*}

\begin{table*}
    \centering
    \setlength{\tabcolsep}{15pt}
    \resizebox{1.0\linewidth}{!}{
    \begin{tabular}{p{.5\linewidth}|p{.5\linewidth}}
        \toprule 
        Query Sentence & Retrieved Videos (illustrated by their corresponding subtitles) \\
        \midrule
        Well, I'm a scientist. & Yeah, there really is. (0.78)	\\
        & I've invited a top \textcolor{blue}{scientist} from Exeter University to the farm. (0.74) \\
        & So, it's an astonishing \textcolor{blue}{discovery}, there's no doubt of that. (0.73) \\
        & \textbf{Well, I'm a scientist. (0.73)} \\
        & There is no miraculous \textcolor{blue}{science} at work here. (0.72) \\
        \midrule 
        She was deeply involved in farming, and her impassioned campaigning certainly ruffled a few feathers.	& \textbf{She was deeply involved in farming, and her impassioned campaigning certainly ruffled a few feathers. (0.74)} \\
        & I have always thought it somewhat odd that the lady, who has a perfectly competent husband, should insist on \textcolor{blue}{managing every detail of farms and woodland problems herself}. (0.66) \\
        & It's \textcolor{blue}{driving} me absolutely \textcolor{blue}{crazy}. (0.66) \\
        & To encourage them to do just that, \textcolor{blue}{farmers} have been \textcolor{blue}{campaigning hard}. (0.65)	\\
        & Yes, he can solve civil engineering problems, mining quarrying problems, but this is \textcolor{blue}{farming}. (0.65) \\
        \midrule
        They had chosen the patch of sky, the Southern Hole, precisely because it was relatively clear of galactic dust.	& So as far as we can tell, the cosmic microwave background looks much the same anywhere that we can observe it on \textcolor{blue}{the sky}, and so the best place to observe it, to look for very faint signals, is where our own \textcolor{blue}{galaxy} and the emission from our galaxy is the faintest. (0.69) \\
        & It's really important, because the telescopes are searching for exceedingly faint signals, signals that might arise from gravitational waves in the early universe. (0.68) \\
        & \textbf{They had chosen the patch of sky, the Southern Hole, precisely because it was relatively clear of galactic dust. (0.67)} \\
        & John and his team have chosen a \textcolor{blue}{particular patch of sky }to search for evidence of gravitational waves. (0.67)	\\
        & In fact, it was impossibly easy to attract the birds. (0.66) \\
        \midrule
        Now, I've got nice... marinana... marinara  & I'm going meatballs in \textcolor{blue}{marinara}. (0.70) \\
        meatballs. & Yeah, mushroomy, yeah (0.69)	\\
        & It's not mushroom coloured, it's nude! (0.68)	\\
        & I'm going with the chicken pesto pasta. (0.68) \\
        & \textbf{Now, I've got nice... marinana... marinara meatballs.} (0.67) \\
        \midrule
        Now, making bread is fairly easy, because you need very few ingredients. & \textbf{Now, making bread is fairly easy, because you need very few ingredients. (0.78)} \\
        & It's this change which \textcolor{blue}{makes bread possible}. (0.77) \\
        & Experts decided they could reinvent \textcolor{blue}{bread}. (0.76)	\\
        & Why does supermarket \textcolor{blue}{bread} stay soft, while home baked \textcolor{blue}{bread }goes stale?  (0.74) \\
        & For home baked \textcolor{blue}{bread you only need flour, water and yeast}. (0.74) \\
\bottomrule
    \end{tabular}
    }
    \vspace{-0.3cm}
    \caption{\textbf{Qualitative T2V retrieval examples} obtained on the unseen \testManual sentences. For visibility, we report the query sentence along with the subtitles corresponding to the retrieved sign videos, as well as their similarities (in the range 0-1). Words in the retrieved subtitle that are identical or 
    semantically similar to those in the query are shown in blue. Note the table continues on the next page.
    }
    \label{tab:app:t2v}
\end{table*}

\begin{table*}
    \centering
    \setlength{\tabcolsep}{15pt}
    \resizebox{1.0\linewidth}{!}{
    \begin{tabular}{p{.5\linewidth}|p{.5\linewidth}}
        \toprule 
        Query Sentences & Retrieved Videos (illustrated by their corresponding subtitles) \\
        \midrule
        His love of the Orient inspired him to transform Batsford, tearing out formal beds in favour of wild planting and exotic trees. & \textbf{His love of the Orient inspired him to transform Batsford, tearing out formal beds in favour of wild planting and exotic trees. (0.67)} \\
        & \textcolor{blue}{The oriental plants and water feature} here were the brainchild of Victorian diplomat Lord Redesdale. (0.66) \\
        & Lord Redesdale converted to Buddhism and hidden \textcolor{blue}{amongst the trees} are Buddhist-themed bronzes, a Japanese bridge and a peace pavilion that reflected \textcolor{blue}{his love} of the culture. (0.66) \\
        & I'm rooted to the ground at \textcolor{blue}{Batsford} Arboretum, home to a wide variety of \textcolor{blue}{unusual tree species from around the world}. (0.66) \\
        & When you wander around a fair like this, one that is so focused on game hunting, you don't naturally think about \textcolor{blue}{wildlife conservation}, but I've been told that the two can go hand-in-hand. (0.64) \\
        \midrule
        An archipelago of 18 islands, 200 miles north west of Shetland, the Faroe Islands are a self-governing nation within the Kingdom of Denmark. & \textbf{An archipelago of 18 islands, 200 miles north west of Shetland, the Faroe Islands are a self-governing nation within the Kingdom of Denmark. (0.75)} \\
        & The \textcolor{blue}{Isle} of Mull... (0.75) \\
        & We're on the Falkland \textcolor{blue}{Islands}... (0.72) \\
        & Yes, we've got Little \textcolor{blue}{Eye} just down there, Middle \textcolor{blue}{Eye} in front of us, and Hilbre just there (0.71) \\
        & Easter \textcolor{blue}{island} was once covered in forest. (0.71) \\
        \midrule
        We're growing tomatoes almost all year round now. & \textbf{We're growing tomatoes almost all year round now. (0.81)} \\
        & To get the best performance from the fruit and the plant, we need to \textcolor{blue}{give the plant} everything that it needs. (0.78) \\
        & Because you need extra heat to \textcolor{blue}{grow tomatoes} out of season, it makes them more expensive. (0.78) \\
        & July to October were always the months to eat \textcolor{blue}{tomatoes} but now vast heated greenhouses mean we can \textcolor{blue}{grow them between February and November}. (0.76)	\\
        & So, why did you choose \textcolor{blue}{tomatoes} as the main medium for this gene? (0.75) \\
        \midrule
        It's size and quality suggests Maryport was a cult centre, drawing people from far and wide to  & \textcolor{blue}{People came to} Lyme Regis to go fossil hunting with \textcolor{blue}{Mary} Anning. (0.69)	\\
       worship. & \textbf{It's size and quality suggests Maryport was a cult centre, drawing people from far and wide to worship. (0.68)} \\
        & This custom is controversial and \textcolor{blue}{draws} objections from the \textcolor{blue}{wider world}. (0.66) \\
        & But inside the earthworm this activity is \textcolor{blue}{magnified} to levels that are truly mind-blowing. (0.66)	\\
        & As pagans, the Romans \textcolor{blue}{worshipped} many gods and spirits. (0.65) \\
        \midrule
        Not just the skeleton, but all the soft tissue, you know, all the muscles and the brain and the fur. & \textbf{Not just the skeleton, but all the soft tissue, you know, all the muscles and the brain and the fur. (0.80)} \\
        & We very rarely get \textcolor{blue}{soft tissue} preservation of extinct animals. (0.70) \\
        & We're concerned that there might be a \textcolor{blue}{bone} infection. (0.69) \\
        & It can insert that into the aphid and it'll suck all the goodness out. (0.68)	\\
        & They're covered in scratches and scars, they're covered in notches, they're covered in \textcolor{blue}{tooth} rakes. (0.68) \\
\bottomrule
    \end{tabular}
    }
    \label{tab:app:t2v_2}
\end{table*}

\end{document}